\documentclass[10pt]{article}

\usepackage{graphicx}
\usepackage{float}
\usepackage{amsmath}
\usepackage{amssymb}
\usepackage{mathtools}
\usepackage{amsthm}
\usepackage{booktabs}
\usepackage[margin=1.5in]{geometry}
\usepackage[hidelinks]{hyperref}
\usepackage{natbib}
\theoremstyle{plain}
\newtheorem{theorem}{Theorem}[section]

\newtheorem{lemma}[theorem]{Lemma}
\newtheorem{corollary}[theorem]{Corollary}
\theoremstyle{definition}

\theoremstyle{definition}
\newtheorem{remark}[theorem]{Remark}

\title{{\Large\bfseries
Gating Enables Curvature: A Geometric Expressivity Gap in Attention
}}

\author{
Satwik Bathula, Anand A. Joshi \\[0.5em]
Department of Electrical and Computer Engineering\\
University of Southern California \\[0.25em]
\texttt{sbathula@usc.edu, ajoshi@usc.edu}
}

\date{}
\begin{document}

\maketitle

\begin{abstract}
 Multiplicative gating is widely used in neural architectures and has recently been applied to attention layers to improve performance and training stability in large language models. Despite the success of gated attention, the mathematical implications of gated attention mechanisms remain poorly understood. We study attention through the geometry of its representations by modeling outputs as mean parameters of Gaussian distributions and analyzing the induced Fisher--Rao geometry. We show that ungated attention operator is restricted to intrinsically flat statistical manifolds due to its affine structure, while multiplicative gating enables non-flat geometries, including positively curved manifolds that are unattainable in the ungated setting. These results establish a geometric expressivity gap between ungated and gated attention. Empirically, we show that gated models exhibit higher representation curvature and improved performance on tasks requiring nonlinear decision boundaries whereas they provide no consistent advantage on tasks with linear decision boundaries. Furthermore, we identify a structured regime in which curvature accumulates under composition, yielding a systematic depth amplification effect.

\end{abstract}


\section{Introduction}

Attention mechanisms are a core part of transformer models used for sequence modeling and large language models \citep{vaswani2017attention}. Prior work studies the expressivity of attention mechanisms, including universality and depth separation \citep{yun2020transformers, NEURIPS2020_ff4dfdf5,pérez2018on,wang2024understanding}. These works study what input-output mappings attention models can represent, independent of training.

Yet, the intrinsic geometric structure of the representation spaces induced by attention layers remains largely unexplored. An attention layer defines a family of representations rather than a single mapping. Understanding the geometry of this family provides a complementary notion of expressivity that focuses on how representations are organized internally rather than which functions can be computed.

We study the geometry induced by attention layers by modeling their outputs as parameters of a statistical manifold equipped with the Fisher--Rao metric, whose curvature captures intrinsic properties of the representation space \citep{amari1985differential,amari2000methods,amari2016information,pmlr-v89-liang19a,pmlr-v89-amari19a,pmlr-v162-kim22f}. For Gaussian decoders, the Fisher geometry associated with the mean parameter is Euclidean, so any nonzero curvature arises solely from the structure of the attention mapping itself. This choice isolates the geometric structure induced by the architecture, rather than by the statistical model \citep{rao1945information,amari1985differential}.

Recent architectural developments have introduced attention mechanisms augmented with multiplicative gating, reporting improvements in training stability, scaling behavior, and long-context modeling \citep{qiu2025gated,NEURIPS2024_d3f39e51,cho2014gru,danihelka2016associative,NIPS2016_f69e505b,pmlr-v70-dauphin17a,PhysRevX.12.011011}. While these results establish the practical benefits of gating, a fundamental question remains as to how multiplicative gating affects the intrinsic geometry of representations produced by attention layers. We show that standard ungated attention has a structural limitation, since its outputs are affine combinations of value vectors, which implies that the associated statistical manifolds are intrinsically flat. In contrast, multiplicative gating removes this affine constraint and enables non-flat geometries.

Recent empirical work has identified a key limitation of standard softmax attention: the composition of the value and output projections induces a low-rank linear bottleneck, restricting the expressiveness of attention representations. In particular, the attention output can be rewritten as
\[
o_i^k = \sum_j S_{ij}^k \, X_j (W_V^k W_O^k),
\]
where the effective transformation is governed by the product $W_V^k W_O^k$, whose rank is at most $d_k \ll d_{\text{model}}$ (Eq.~(6), Sec.~4.1 in~\cite{qiu2025gated}). This shows that attention operates through a low-rank linear mapping.

We show that this algebraic constraint has a direct geometric interpretation. Since attention outputs are convex combinations of value vectors, they admit an affine representation whose image lies in a low-dimensional subspace of the ambient space. As we prove, such affine embeddings induce intrinsically flat statistical manifolds. Thus, the low-rank bottleneck can be understood as a restriction to flat representation geometry. Multiplicative gating removes this constraint by introducing input-dependent nonlinear modulation, enabling non-flat (curved) geometries.

In this work, we study gated and ungated attention from a geometric perspective, with the following contributions:
\begin{itemize}
\item We introduce a geometric framework for analyzing attention by modeling outputs as mean parameters of Gaussian distributions and studying the induced statistical manifolds under the Fisher--Rao metric.

\item We show that standard ungated attention is restricted to intrinsically flat statistical manifolds, as a consequence of the output’s affine structure.

\item We provide a geometric interpretation of the low-rank bottleneck in attention by showing that the affine structure of ungated attention confines representations to low-dimensional subspaces, inducing intrinsically flat geometry.

\item We prove that introducing a multiplicative gate strictly enlarges the class of realizable geometries in the sense that it enables statistical manifolds with strictly positive curvature that are provably unattainable by ungated attention.

\item We show that this phenomenon persists in arbitrary output dimensions $D \geq 3$ and within fully content-aware attention architectures.

\item We analyze the resulting geometry under small parameter perturbations, showing that positive curvature is intrinsic and locally stable.

\item We complement the theory with a synthetic experiment showing that gated models can exhibit higher representation curvature than ungated counterparts, illustrating that the geometric flexibility enabled by gating can manifest in trained models.

\item We connect curvature to representation structure by showing that affine attention representations induce flat geometry, while multiplicative gating enables non-affine (curved) representations.
\end{itemize}
\paragraph{Representation-level implication.}
Ungated attention produces representations confined to affine (flat) geometry, which can limit the ability to directly capture non-affine structure at the level of a single block. Multiplicative gating relaxes this constraint, enabling intrinsically curved representations and supporting a richer class of representational geometries, particularly for data with nonlinear or manifold structure.

These results characterize the representational structure induced by a single attention block, providing a pointwise geometric perspective that complements the depth scaling analysis developed later in the paper.

This perspective parallels combinatorial expressivity results for deep networks \cite{NIPS2014_fa6f2a46}, where depth enables growth in representational complexity under suitable structural conditions. In contrast, our results characterize a geometric mechanism through which multiplicative gating can induce and amplify intrinsic curvature.

\begin{figure}[H]
    \centering
    \includegraphics[width=1\linewidth]{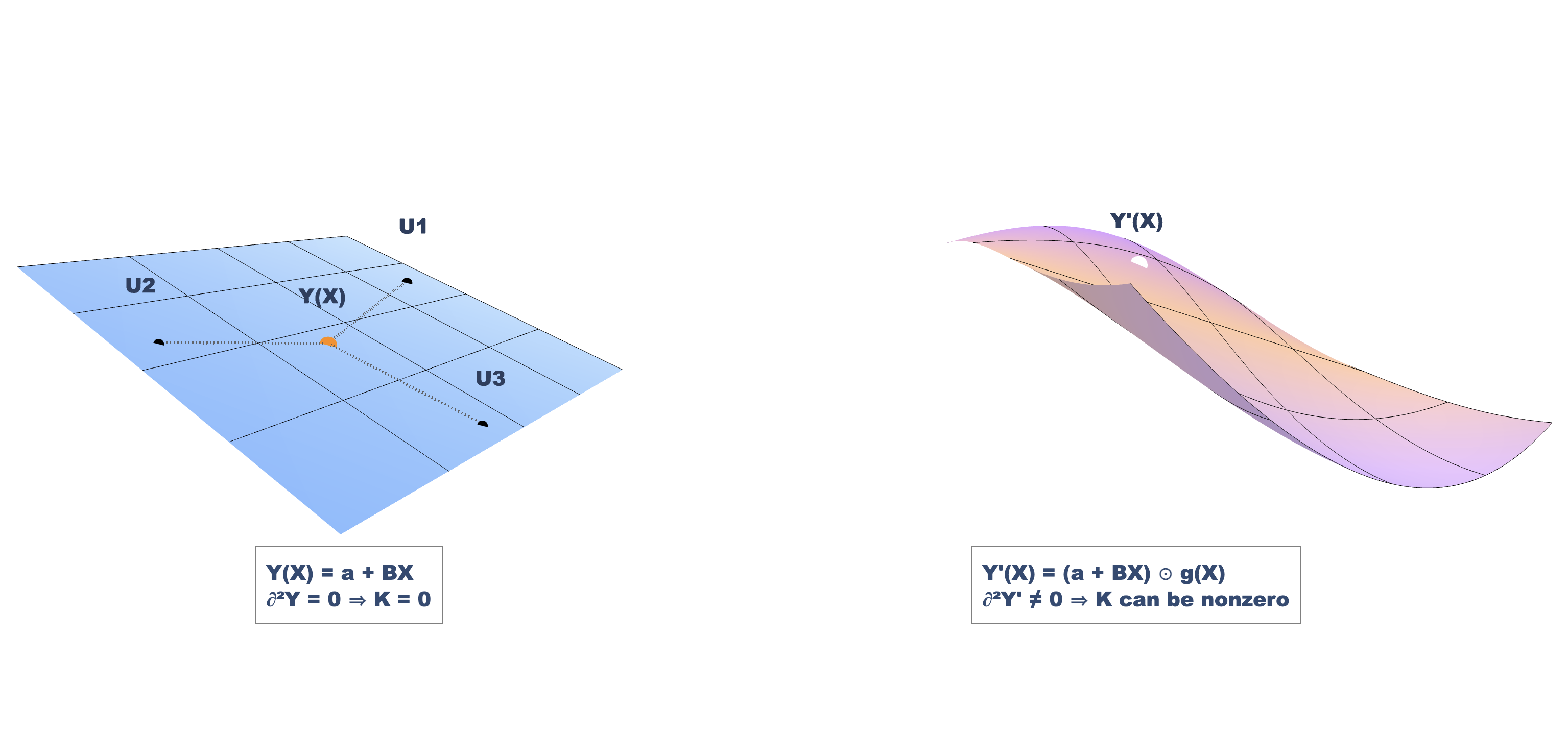}
    \caption{\textbf{ Geometric intuition for curvature generation in attention.}
Left: Ungated attention produces affine combinations of value vectors, so outputs lie in the affine hull $\mathrm{aff}\{U_1,U_2,U_3\}$, yielding a flat representation manifold with zero curvature.
Right: Gating introduces element-wise modulation $Y'(X) = Y(X) \odot g(X)$, breaking affine structure and enabling nonzero curvature in the induced representation manifold.}
    \label{fig:placeholder}
\end{figure}

\section{Preliminaries}

\subsection{Notations}

\paragraph{Linear algebra.}
All vectors are real-valued. We write $\mathbb{R}^D$ for the ambient representation space and use lowercase letters for vectors and uppercase letters for matrices. Inner products and norms are taken with respect to the standard Euclidean structure on $\mathbb{R}^D$. For vectors $a, b \in \mathbb{R}^D$, $a \odot b$ denotes their componentwise (Hadamard) product.

\paragraph{Probability simplex.}
The probability simplex in $\mathbb{R}^n$ is denoted by
\[
\Delta^{n-1} := \left\{ \alpha \in \mathbb{R}^n \,\middle|\, \alpha_i \geq 0, \ \sum_{i=1}^n \alpha_i = 1 \right\}.
\]

\paragraph{Gaussian statistical manifolds.}
We consider Gaussian distributions of the form
\[
p(y \mid \mu) = \mathcal{N}(y; \mu, I_D),
\]
where the mean parameter $\mu \in \mathbb{R}^D$ varies and the covariance is fixed. This defines a statistical manifold parameterized by $\mu$.

\paragraph{Fisher--Rao geometry and fixed covariance.}
The Fisher--Rao metric provides a canonical, parameterization-invariant notion of geometry for statistical models, ensuring that geometric quantities such as curvature are intrinsic and not dependent on the choice of coordinates. For a Gaussian location family with fixed covariance $\Sigma \succ 0$,
\[
p(y \mid \mu) = \mathcal{N}(y; \mu, \Sigma),
\]
the Fisher information for the mean parameter is given by
\[
I(\mu) = \Sigma^{-1}.
\]
Consequently, for any parameterization $\mu = \mu(\phi)$, the induced Riemannian metric is
\[
g_{ij}(\phi) = \partial_i \mu(\phi)^\top \Sigma^{-1} \partial_j \mu(\phi),
\]
which corresponds to a Mahalanobis inner product. Defining the whitened mean $\tilde{\mu}(\phi) = \Sigma^{-1/2} \mu(\phi)$, this metric reduces to the Euclidean form
\[
g_{ij}(\phi) = \langle \partial_i \tilde{\mu}(\phi), \partial_j \tilde{\mu}(\phi) \rangle,
\]
showing that any fixed non-isotropic covariance is equivalent to the isotropic case under a linear change of coordinates.

As a result, the intrinsic geometry of any statistical manifold induced by $\mu(\phi)$ is determined entirely by the embedding $\phi \mapsto \mu(\phi) \subset \mathbb{R}^D$. All geometric properties therefore depend only on $\mu(\phi)$. For Gaussian location families, the Fisher--Rao metric coincides with the Euclidean metric and serves only as a coordinate-invariant formulation of the induced geometry. In particular, since the metric is constant, any nonzero curvature arises from the embedding $\mu(\phi)$ rather than from the statistical model. If the covariance were allowed to depend on the input, then the Fisher metric itself would vary with $\phi$ and contribute additional curvature, even when $\mu(\phi)$ is affine, making it difficult to attribute geometric effects to the attention mechanism alone. This choice therefore provides a minimal setting in which any observed curvature is attributable solely to the attention mapping.

We consider a Gaussian location family with fixed covariance, under which the Fisher–Rao metric on the mean parameter reduces to a constant (Euclidean) metric. Consequently, the intrinsic geometry of the induced statistical manifold is determined entirely by the embedding $\mu(\phi) \subset \mathbb{R}^D$, and any nonzero curvature arises solely from the structure of this mapping.

This construction should be viewed as a coordinate-invariant realization of a constant Riemannian metric on the representation space, rather than as a modeling assumption on the data distribution. In more general likelihood models where the metric varies with the parameters, additional curvature may arise from the metric itself. Our formulation isolates the geometric contribution of the attention operator.

\paragraph{Curvature.}
We use standard notions from Riemannian geometry. The Riemann curvature tensor is denoted by $R$, and sectional curvature by $K$. A manifold is said to be flat if $R \equiv 0$, and to have positive curvature if sectional curvatures are strictly positive on an open set.

\paragraph{Attention outputs.}
Throughout the paper, attention outputs are denoted by $Y(X) \in \mathbb{R}^D$ and are interpreted as mean parameters of Gaussian distributions. Ungated attention outputs take the form
\[
Y(X) = \sum_{i=1}^n \alpha_i(X) U_i, \quad \alpha(X) \in \Delta^{n-1},
\]
where $U_i \in \mathbb{R}^D$ are value vectors.

Although the attention weights $\alpha_i(X)$ depend nonlinearly on the input through the softmax, the output $Y(X)$ remains an affine combination of the value vectors. It is this affine structure of the output, rather than the nonlinearity in the weights, that determines the geometry of the induced representation manifold.

Gated attention outputs are of the form
\[
Y'(X) = Y(X) \odot \sigma(X_g(X) W_\theta),
\]
where $\sigma$ is a smooth elementwise nonlinearity with values in $(0,1)$, $X_g(X)$ denotes a gating input, and $W_\theta$ is a trainable matrix.

Unless stated otherwise, all geometric statements are local and hold on open sets where the parameterization $\mu(\phi)$ of the statistical manifold is smooth and has full-rank Jacobian.

\subsection{Problem Setup and Formulation}

We formalize attention architectures and the geometric objects they induce. Our goal is to compare the intrinsic geometry of representations produced by ungated and gated attention mechanisms, independent of training dynamics and downstream tasks.

\paragraph{Attention architecture.}
We consider a single-head attention layer of the form introduced by \cite{vaswani2017attention}. Let $X$ denote an input sequence, and let
\[
U_1, \dots, U_n \in \mathbb{R}^D
\]
be value vectors obtained after value and output projections. The ungated attention output is given by
\[
Y(X) = \sum_{i=1}^n \alpha_i(X) U_i, \quad \alpha(X) \in \Delta^{n-1},
\]
where the attention weights $\alpha_i(X)$ are produced by a softmax over query--key interactions.

For fixed value vectors $U_1, \dots, U_n$, the mapping $X \mapsto Y(X)$ takes values in the affine span of $\{U_1, \dots, U_n\}$, reflecting the affine structure of ungated attention outputs.

\paragraph{Gated attention.}
We consider attention augmented with a multiplicative gate. Let $X_g(X) \in \mathbb{R}^m$ denote a gating input, $W_\theta \in \mathbb{R}^{m \times D}$ a trainable matrix, and $\sigma : \mathbb{R} \to (0,1)$ a smooth elementwise nonlinearity. The gated attention output is
\[
Y'(X) = Y(X) \odot \sigma(X_g(X) W_\theta).
\]
When the gate is constant, this reduces to the ungated case up to a fixed rescaling. Allowing the gate to vary introduces nonlinear modulation of the attention output.

\paragraph{Representation geometry.}
To study the geometry of attention representations, we associate each output with a Gaussian distribution whose mean is given by the attention output,
\[
p(y \mid \mu) = \mathcal{N}(y; \mu, I_D), \quad \mu(X) = Y(X).
\]
As discussed in Section~2.1, the Fisher--Rao metric for this model is Euclidean on the mean parameter, so the induced geometry is determined entirely by the embedding $\mu(X) \subset \mathbb{R}^D$. In particular, curvature reflects nonlinear structure in the mapping $X \mapsto Y(X)$.

This provides a principled way to study the intrinsic geometry of representations produced by attention layers, independent of parameterization.

\paragraph{Problem formulation.}
We compare the statistical manifolds induced by ungated and gated attention,
\[
\begin{aligned}
\mathcal{M}_{\mathrm{ung}} &:= \{ \mathcal{N}(Y(X), I_D) \mid X \in \mathcal{X} \}, \\
\mathcal{M}_{\mathrm{gat}} &:= \{ \mathcal{N}(Y'(X), I_D) \mid X \in \mathcal{X} \}.
\end{aligned}
\]

Our goal is to characterize the geometric constraints imposed by the affine structure of ungated attention, and to determine whether multiplicative gating enlarges the class of realizable geometries. In particular, we ask:
\begin{itemize}
    \item What geometric structures are imposed by the affine form of ungated attention?
    \item Can multiplicative gating enable representations with nonzero intrinsic curvature?
    \item Does gating strictly enlarge the range of geometries realizable by attention?
\end{itemize}

\section{Main Results}

This section develops the main theoretical results describing the intrinsic geometry of statistical manifolds induced by ungated and gated attention operators. Proofs of all results in this section are deferred to the appendix.

We first examine the intrinsic geometry induced by ungated attention under the Gaussian decoder.

\begin{theorem}[Flatness of ungated attention manifolds]
An ungated attention operator whose coefficients range over the relative interior of the simplex and whose outputs are affine combinations of value vectors and hence lie in an affine subspace, equipped with a Gaussian decoder, induces an intrinsically flat statistical manifold.
\end{theorem}

More precisely, let $U_1, \dots, U_n \in \mathbb{R}^D$ be fixed value vectors and define the convex hull
\[
\mathcal{U} := \mathrm{conv}(U_1, \dots, U_n) = \left\{ \sum_{i=1}^n \alpha_i U_i \,\middle|\, \alpha \in \Delta_{n-1} \right\} \subset \mathbb{R}^D.
\]
Let $k := \dim \operatorname{aff}\{U_1, \ldots, U_n\}
= \dim \operatorname{span}\{U_1 - U_n, \ldots, U_{n-1} - U_n\}.$ On the relative interior $\operatorname{relint}(\mathcal{U})$, choose affine coordinates
\[
\begin{aligned}
\varphi &: \Omega \subset \mathbb{R}^k \to \operatorname{relint}(\mathcal{U}), \\
\mu_{\mathrm{ung}}(\phi) &:= \varphi(\phi) = a + B\phi, \quad \text{for } \phi \in \Omega.
\end{aligned}
\]
where $B \in \mathbb{R}^{D \times k}$ has full column rank. Consider the statistical manifold
\[
\mathcal{M}_{\mathrm{ung}} := \{ \mathcal{N}(\mu_{\mathrm{ung}}(\phi), I_D) \mid \phi \in \Omega \},
\]
equipped with the Fisher--Rao metric. Then the induced Riemannian metric on $\Omega$ is constant:
\[
g^{\mathrm{ung}}_{ij}(\phi) = \langle \partial_i \mu_{\mathrm{ung}}(\phi), \partial_j \mu_{\mathrm{ung}}(\phi) \rangle = C_{ij},
\]
where $C = (C_{ij})$ is a fixed positive-definite matrix. Consequently, the Levi--Civita connection has vanishing Christoffel symbols, and the Riemann curvature tensor and sectional curvatures satisfy
\[
R^{\mathrm{ung}} \equiv 0, \qquad K_{\mathrm{ung}} = 0 \quad \text{on } \Omega.
\]

Thus, $\mathcal{M}_{\mathrm{ung}}$ is isometric to an open subset of the Euclidean space $(\mathbb{R}^k, C)$ and is intrinsically flat.

\begin{remark}[Intrinsic flatness of ungated attention]
The flatness established in Theorem~3.1 is intrinsic to ungated attention and does not depend on the chosen parameterization. In affine coordinates $\phi \in \Omega \subset \mathbb{R}^k$, the induced Fisher--Rao metric satisfies
\[
g^{\mathrm{ung}}_{ij}(\phi) = \langle \partial_i \mu_{\mathrm{ung}}(\phi), \partial_j \mu_{\mathrm{ung}}(\phi) \rangle = C_{ij},
\]
for a constant positive-definite matrix $C$. Hence the Levi--Civita connection is flat and the Riemann curvature tensor vanishes identically on $\Omega$.

For any diffeomorphism $\psi : \hat{\Omega} \to \Omega$, the pullback metric $\tilde{g} = \psi^* g^{\mathrm{ung}}$ satisfies
\[
R(\tilde{g}) = \psi^* R(g^{\mathrm{ung}}) \equiv 0.
\]

Therefore no smooth reparameterization can induce nonzero curvature.

The vanishing curvature therefore shows a fundamental geometric limitation of ungated attention, arising from the affine form of its output together with the Euclidean Fisher geometry of the Gaussian mean family.
\end{remark}

The above result also admits an interpretation in terms of the low-rank structure of attention.

\begin{remark}[Connection to low-rank bottleneck]
The affine form \( \mu_{\mathrm{ung}}(\phi) = a + B\phi \) reflects the fact that attention outputs are convex combinations of value vectors, arising from the simplex constraint \( \alpha(X) \in \Delta^{n-1} \). Consequently, the output lies in the affine hull of these vectors. The effective dimensionality of this subspace is governed by the span of the value matrix, and in particular by the rank of \( B \in \mathbb{R}^{D \times k} \), whose columns span a \( k \)-dimensional subspace. Thus, \( \mu_{\mathrm{ung}}(\phi) \) takes values in the affine subspace \( a + \mathrm{span}(B) \), and although \( \mu_{\mathrm{ung}}(\phi) \in \mathbb{R}^D \), its image is confined to a \( k \)-dimensional affine subspace.

This aligns with prior observations that attention induces a low-rank linear mapping through the value projection \( V = X W_V \) followed by the output mapping \( W_O \), whose rank is bounded by $d_k \ll d_{\mathrm{model}}$ (see Eq.~(6), Sec.~4.1 in~\cite{qiu2025gated}). Although the bound $k \le n - 1$ is structural, in typical transformer architectures the effective dimension is much smaller than the ambient dimension, so that $k \ll D$ in practice. In our formulation, this affine structure induces a constant metric and hence intrinsic flatness of the representation manifold.

\end{remark}

In this sense, the low-rank bottleneck is not merely an algebraic property of the attention operator, but a geometric one, corresponding to an intrinsic restriction to flat representation manifolds.

We next consider the effect of introducing a multiplicative gate into the attention output.

\begin{lemma}[Multiplicative gating generically destroys affine structure]
Let $U\subset \mathbb{R}^d$ be a connected open set with $d\ge 1$. Let
\[
Y(\phi)=a+B\phi,\qquad B\in \mathbb{R}^{D\times d},\ \mathrm{rank}(B)=d,
\]
and denote by $B_i\in\mathbb{R}^D$ the $i$-th column of $B$. For $g\in C^2(U,\mathbb{R}^D)$ define
\[
\mu(\phi)=Y(\phi)\odot g(\phi),
\]
where $\odot$ denotes the componentwise (Hadamard) product. Then for all $i,j$,
\[
\partial_{ij}\mu
=
B_i\odot \partial_j g
+
B_j\odot \partial_i g
+
Y\odot \partial_{ij}g.
\]
Moreover, the set
\[
\mathcal A=\{g\in C^2(U,\mathbb{R}^D): \mu \text{ is affine on } U\}
\]
is a proper closed linear subspace of $C^2(U,\mathbb{R}^D)$ (with its Fréchet topology). Hence $\mathcal A$ is nowhere dense and meagre, and for residual $g\in C^2(U,\mathbb{R}^D)$, the map $\mu$ is non-affine. Equivalently, non-affinity holds generically in the Baire category sense.
\end{lemma}

\paragraph{From generic non-affinity to geometric flexibility.}
Lemma~3.4 shows that multiplicative gating generically destroys the affine structure of the embedding $\mu(\phi) = Y(\phi) \odot g(\phi)$. In particular, for generic $g$, the second derivative $D^2 \mu$ does not vanish identically and the Jacobian $J_\mu$ is non-constant on open sets. Since the induced Fisher--Rao metric is $G(\phi) = J_\mu(\phi)^\top J_\mu(\phi)$, this yields a non-constant metric. A non-constant metric does not in general imply nonzero curvature, but it reflects a departure from affine structure. Establishing nonzero curvature requires either an explicit curvature computation or a construction with known curvature. This is formalized below by explicit constructions and genericity results.

\paragraph{Existence of curved realizations (witness construction).}
While Lemma~3.4 establishes that multiplicative gating breaks affine structure, it does not specify the types of geometries that can be realized. The following theorem provides a concrete witness, showing that multiplicative gating can realize strictly positively curved statistical manifolds.The broader structural result is that flatness is nongeneric under multiplicative gating.

\begin{theorem}[Concrete witness of non-flat geometry under multiplicative gating]\label{thm:positive_curvature_gating}
There exist
\begin{itemize}
    \item an output dimension $D=3$,
    \item a connected open set $U\subset \mathbb{R}^2$,
    \item a smooth affine map $Y:U\to \mathbb{R}^3$,
    \item a smooth map $X:U\to \mathbb{R}$,
    \item and a matrix $W_\theta\in \mathbb{R}^{1\times 3}$ with $\operatorname{rank}(W_\theta)=1$,
\end{itemize}
such that the following hold:
\begin{enumerate}
    \item[(i)] the ungated map
    \[
    \mu_{\mathrm{ung}}(\phi):=Y(\phi)
    \]
    is affine, and the corresponding statistical manifold induced under the Gaussian decoder
    \[
    p(y\mid \mu)=\mathcal N(y;\mu,I_3)
    \]
    is intrinsically flat.
    \item[(ii)] the gated map
    \[
    \mu_{\mathrm{gat}}(\phi)
    :=
    Y(\phi)\odot \sigma\!\bigl(X(\phi)W_\theta\bigr)
    \]
    parametrizes a smooth patch of the unit sphere $S^2\subset \mathbb{R}^3$. Consequently, the induced Fisher--Rao metric has Gaussian curvature
    \[
    K_{\mathrm{gat}}(\phi)\equiv 1>0
    \qquad \text{for all }\phi\in U.
    \]
\end{enumerate}
\end{theorem}

\begin{corollary}[Realizability in gated attention]\label{cor:attention_realization}
There exists a single-head, single-layer attention block with multiplicative gating of the form
\[
Y' = Y \odot \sigma(XW_\theta),
\]
such that, under the Gaussian decoder $p(y\mid \mu)=\mathcal{N}(y;\mu,I_D)$, the induced statistical manifolds satisfy
\[
K_{\mathrm{ung}} \equiv 0,
\qquad
K_{\mathrm{gat}}(\phi) \equiv 1 > 0 \quad \text{for all } \phi \in U.
\]

In particular, multiplicative gating in attention architectures can realize strictly positively curved statistical manifolds that ungated attention cannot realize.
\end{corollary}

\paragraph{Curvature gap inside the architectural class.}
Lemma~3.4 establishes that multiplicative gating generically destroys affine structure, while Theorem~3.5 provides an explicit curved witness in an abstract gated embedding. The next theorem strengthens this further by showing that the same geometric separation can be realized inside a standard content-aware transformer attention block. In this sense, the curvature gap is not merely a property of an abstract construction, but an expressivity phenomenon already present within the architectural class itself.

\begin{theorem}[Existence of a curvature gap inside content-aware attention]\label{thm:content_aware_gap}
Consider a single-head, single-layer transformer attention block with standard content-aware projections
\[
\begin{aligned}
Q(X) &= XW_Q, \\
K(X) &= XW_K, \\
V(X) &= XW_V, \\
U(X) &= V(X)W_O.
\end{aligned}
\]
We adopt the convention that all vectors are represented as row vectors, and linear maps act on the right.

For a fixed position $j$, let
\[
Y_j(X)=\sum_{i=1}^n \alpha_{ji}(X)\,U_i(X)
\]
denote the attention output at position $j$, where the weights $\alpha_{ji}(X)$ are obtained by applying the softmax to $Q(X)K(X)^\top$.
Let the decoder be the Gaussian family
\[
p(y\mid \mu)=\mathcal N(y;\mu,I_3),
\]
so that the Fisher--Rao metric on the mean parameter is Euclidean.
Define the gated output at position $j$ by
\[
Y'_j(X) = Y_j(X) \odot \sigma(X_g(X) W_\theta),
\]
where $X_g$ is a differentiable function of $X$, $W_\theta\in\mathbb R^{m\times 3}$ has rank $3$, and $\sigma:(-\infty,\infty)\to(0,1)$ is a smooth elementwise nonlinearity.

Then there exist
\begin{itemize}
    \item a connected open set $U\subset \mathbb R^2$,
    \item a smooth map $\phi\mapsto X(\phi)$ producing input sequences,
    \item projection matrices $W_Q,W_K,W_V,W_O$,
    \item and a gating matrix $W_\theta\in\mathbb R^{m\times 3}$ with $\operatorname{rank}(W_\theta)=3$,
\end{itemize}
such that the following hold simultaneously.

\begin{enumerate}
    \item[(i)] \textbf{The ungated output realizes an affine (flat) map in $\phi$.}
    For all $\phi\in U$,
    \[
    Y_j(X(\phi))=a+B\phi
    \]
    for some $a\in\mathbb R^3$ and full-rank $B\in\mathbb R^{3\times 2}$.
    The induced statistical manifold
    \[
    \mathcal M_{\mathrm{ung}}
    :=
    \{\mathcal N(Y_j(X(\phi)),I_3):\phi\in U\}
    \]
    is isometric to an open subset of Euclidean space, and hence
    \[
    K_{\mathrm{ung}}(\phi)=0
    \qquad
    \text{for all }\phi\in U.
    \]

    \item[(ii)] \textbf{Gated output is spherical.}
    There exists a differentiable choice of gating input $X_g(\phi)$ such that
    \[
    Y'_j(X(\phi))=s(\phi),
    \qquad
    s(\phi)=
    \begin{pmatrix}
    \cos\phi_1\cos\phi_2\\
    \cos\phi_1\sin\phi_2\\
    \sin\phi_1
    \end{pmatrix},
    \]
    which parametrizes a smooth patch of the unit sphere $S^2\subset\mathbb R^3$.
    The induced statistical manifold
    \[
    \mathcal M_{\mathrm{gat}}
    :=
    \{\mathcal N(Y'_j(X(\phi)),I_3):\phi\in U\}
    \]
    is isometric to a spherical patch and satisfies
    \[
    K_{\mathrm{gat}}(\phi)=1
    \qquad
    \text{for all }\phi\in U.
    \]
\end{enumerate}

Thus multiplicative gating introduces a strictly richer geometric class within standard content-aware attention. The construction below provides an explicit witness to the curvature gap. Ungated attention remains flat, whereas gated attention realizes a strictly positively curved statistical manifold inside the same architectural family, where ungated attention is a special case of gated attention.
\end{theorem}

\begin{corollary}[Geometric expressivity of the multiplicative gate]
Fix the projection matrices $W_Q, W_K, W_V, W_O$ and the input map $\phi \mapsto X(\phi)$ constructed in Theorem~3.7. Let the only trainable parameter be the gating matrix $W_\theta$ in
\[
Y'_j(X) = Y_j(X) \odot \sigma\bigl(X_g(X)\, W_\theta\bigr).
\]

Then the family of statistical manifolds realizable by varying $W_\theta$ strictly contains the ungated family (obtained when the gate is constant).

In particular, there exist choices of $W_\theta$ such that
\[
K_{\mathrm{ung}} \equiv 0,
\qquad
K_{\mathrm{gat}}(\phi) \equiv 1 > 0,
\quad \text{for all } \phi \in U.
\]

Thus, multiplicative gating strictly enlarges the set of realizable statistical manifolds within the same attention architecture.
\end{corollary}

\begin{corollary}[Extension to higher-dimensional decoders]
The construction of Theorem~3.7 extends to Gaussian decoders in dimension $D \ge 3$.

Specifically, embedding the spherical patch $s(\phi)$ in the first three coordinates of $\mathbb{R}^D$ and keeping the remaining coordinates constant yields a gated statistical manifold whose sectional curvatures in planes tangent to the spherical directions are strictly positive, while the ungated manifold remains flat.

Thus, the curvature gap persists for all $D \ge 3$.
\end{corollary}

\subsection*{Depth amplification under composition}

The preceding results establish that multiplicative gating enables intrinsically curved representations within a single attention block, and that this curvature gap persists within standard content-aware architectures. However, these results do not address how such geometric structure behaves under composition across layers. In particular, it is not clear whether curvature introduced by gating remains a local effect or can accumulate with depth. To analyze this, we consider a regime in which the gated attention stack admits a local coordinate representation with coherent multiplicative structure across layers. This leads to a conditional local normal form, under which the geometric effect of gating can be tracked explicitly and shown to amplify under composition.

\begin{lemma}[Local gated-attention normal form]
Let $F_L : \mathcal X \to \mathbb R^D$ be the representation map of an $L$-layer gated attention stack. Suppose there exist a point $x_0 \in \mathcal X$, an open neighborhood $x_0 \in U$, a local input chart $\chi_{\mathrm{in}} : U \to V \subset \mathbb R^2$, and a local output chart $\chi_{\mathrm{out}} : F_L(U) \to W \subset \mathbb R^D$ such that
\[
\chi_{\mathrm{in}}(x_0)=0
\]
and
\[
\chi_{\mathrm{out}} \circ F_L \circ \chi_{\mathrm{in}}^{-1}(u,v)
=
\Bigl(u,\ v,\ \sum_{\ell=1}^L a_\ell \psi(u,v),\ 0,\dots,0\Bigr),
\]
for some coefficients $a_\ell > 0$ and some $\psi \in C^2(V)$.

Define
\[
A_L := \sum_{\ell=1}^L a_\ell.
\]
Then
\[
\chi_{\mathrm{out}} \circ F_L \circ \chi_{\mathrm{in}}^{-1}(u,v)
=
\bigl(u,\ v,\ A_L \psi(u,v),\ 0,\dots,0\bigr).
\]
\end{lemma}

The representation in Lemma 3.10 should be interpreted as a local structural reduction rather than a universal representation theorem for gated attention. It isolates a regime in which multiplicative gating induces coherent additive accumulation across layers. The results that follow characterize the geometric consequences of this mechanism, rather than asserting that arbitrary gated attention models satisfy this form. Such conditional structural analyses are standard in the theory of deep networks, where results are often established under specific parameter regimes to isolate mechanisms of expressivity.

\paragraph{Realizability of the normal form.}
The local normal form in Lemma 3.10 is not merely hypothetical. Using constructions similar to Theorem 3.7, one can explicitly realize gated attention stacks in which the attention weights are uniform and the value and output projections align the ungated output with a fixed coordinate direction. Choosing gating inputs so that each layer produces a scalar modulation of the form $a_\ell \psi(u,v)$, the residual connection accumulates these contributions across layers. In such a regime, the representation map reduces locally to the stated normal form.

\begin{lemma}[Realizability of the local gated-attention normal form]
Fix an integer $L \ge 1$ and positive coefficients $a_1,\dots,a_L$.
Let $U \subset \mathbb R^2$ be a connected open set, and let $\psi \in C^2(U)$ satisfy
\[
0 < a_\ell \psi(u,v) < 1
\qquad
\text{for all } (u,v)\in U \text{ and all } \ell=1,\dots,L.
\]
Then there exists an $L$-layer gated attention stack with multiplicative gating of the form
\[
Y' = Y \odot \sigma(X_g W_\theta),
\]
together with residual connections and a smooth local input parametrization
\[
(u,v) \mapsto X(u,v),
\]
such that the induced representation map is exactly
\[
(u,v)\longmapsto \bigl(u,\ v,\ \sum_{\ell=1}^L a_\ell \psi(u,v),\ 0,\dots,0\bigr).
\]
\end{lemma}

\begin{lemma}[Gaussian curvature of the graph normal form]
Let $U \subset \mathbb R^2$ be open with $0 \in U$, and let $\psi \in C^2(U)$ satisfy
\[
\nabla \psi(0)=0.
\]
For $A_L>0$, define
\[
F_L(u,v)=\bigl(u,\ v,\ A_L\psi(u,v),\ 0,\dots,0\bigr)\in\mathbb R^D,
\]
and let
\[
M_L := F_L(U)
\]
with the Riemannian metric induced from the Euclidean metric on $\mathbb R^D$.
Then the Gaussian curvature of $M_L$ at $F_L(0)$ is
\[
K_{M_L}\!\bigl(F_L(0)\bigr)=A_L^2\,\det D^2\psi(0).
\]
\end{lemma}

\begin{theorem}[Depth-amplified intrinsic curvature under a gated-attention normal form]
Under the assumptions of the preceding two lemmas,
\[
K_{M_L}\!\bigl(F_L(0)\bigr)=A_L^2\,\det D^2\psi(0).
\]
In particular, if there exists $a_0>0$ such that $a_\ell \ge a_0$ for all $\ell$, then
\[
\bigl|K_{M_L}\!\bigl(F_L(0)\bigr)\bigr|
\ge
a_0^2\,\bigl|\det D^2\psi(0)\bigr|\,L^2.
\]
If in addition $\det D^2\psi(0)>0$, then
\[
K_{M_L}\!\bigl(F_L(0)\bigr)
\ge
a_0^2\,\det D^2\psi(0)\,L^2.
\]
\end{theorem}

Note that the curvature scaling depends on the cumulative amplitude $A_L = \sum_{\ell=1}^L a_\ell$, and not on depth alone. In particular, a single-layer model with sufficiently large gate amplitude could achieve the same curvature. The result therefore characterizes a regime in which depth provides a systematic mechanism for accumulating curvature, rather than being the only means of achieving it.

\begin{corollary}[Higher-dimensional embedding]
Let $D\ge 3$, and define
\[
\widetilde F_L(u,v)
=
\bigl(u,\ v,\ A_L\psi(u,v),\ 0,\dots,0\bigr)\in\mathbb R^D.
\]
Let $\widetilde M_L := \widetilde F_L(U)$. Then
\[
K_{\widetilde M_L}\!\bigl(\widetilde F_L(0)\bigr)
=
A_L^2\,\det D^2\psi(0),
\]
and consequently
\[
\bigl|K_{\widetilde M_L}\!\bigl(\widetilde F_L(0)\bigr)\bigr|
\ge
a_0^2\,\bigl|\det D^2\psi(0)\bigr|\,L^2
\]
whenever $a_\ell\ge a_0>0$ for all $\ell$. If also $\det D^2\psi(0)>0$, then
\[
K_{\widetilde M_L}\!\bigl(\widetilde F_L(0)\bigr)
\ge
a_0^2\,\det D^2\psi(0)\,L^2.
\]
\end{corollary}

\begin{proof}
The additional coordinates are constant, so they do not affect the induced metric. Thus the intrinsic geometry of $\widetilde M_L$ is identical to that of the graph surface in $\mathbb R^3$. The curvature identity follows immediately from the theorem, and the bounds follow from the same estimate on $A_L$.
\end{proof}

\paragraph{Interpretation.}
Theorem 3.13 shows that, under the local gated-attention normal form, the intrinsic curvature at the base point satisfies
\[
K_{M_L}(F_L(0)) = A_L^2 \det D^2\psi(0), \qquad A_L = \sum_{\ell=1}^L a_\ell.
\]
Thus the geometric effect of multiplicative gating is amplified under composition through the cumulative coefficient \(A_L\). In particular, if there exists \(a_0 > 0\) such that \(a_\ell \ge a_0\) for all \(\ell\), then \(|K_{M_L}(F_L(0))|\) admits a quadratic lower bound in \(L\). This identifies a conditional structural regime in which non-flat geometry does not remain a purely local single-layer phenomenon, but can accumulate coherently across depth. In contrast, in the affine regime identified for ungated attention, the induced representation geometry remains flat, whereas gated attention admits a regime in which curvature grows systematically under composition.

\paragraph{From aligned scalar structure to general vector-valued representations.}
The preceding result characterizes a structured regime in which layer-wise
contributions align along a single direction, yielding coherent quadratic
scaling of curvature. We now show that the underlying geometric mechanism extends to a more general setting,
in which representations are modeled as vector-valued graphs, without requiring alignment across layers.

\begin{theorem}[Curvature of vector-valued graph representations]
Let $U \subset \mathbb{R}^2$ be an open set with $0 \in U$. Let
\[
\Phi_1,\dots,\Phi_L : U \to \mathbb{R}^{D-2}
\]
be $C^2$ maps, and define the aggregated map
\[
\Phi_L^{\mathrm{tot}}(u,v) := \sum_{\ell=1}^L \Phi_\ell(u,v).
\]
Define the immersion
\[
F_L(u,v) := \bigl(u,v,\Phi_L^{\mathrm{tot}}(u,v)\bigr) \in \mathbb{R}^D,
\]
and let $M_L := F_L(U)$ be equipped with the metric induced from the Euclidean
metric on $\mathbb{R}^D$.

Assume that $D\Phi_L^{\mathrm{tot}}(0)=0$. Then the Gaussian curvature of $M_L$
at the point $F_L(0)$ is given by
\[
K_{M_L}\bigl(F_L(0)\bigr)
=
\sum_{\alpha=1}^{D-2}
\det\!\Bigl(D^2(\Phi_L^{\mathrm{tot}})^\alpha(0)\Bigr).
\]
Equivalently,
\[
K_{M_L}\bigl(F_L(0)\bigr)
=
\sum_{\alpha=1}^{D-2}
\det\!\left(
\sum_{\ell=1}^L D^2 \Phi_\ell^\alpha(0)
\right).
\]
\end{theorem}

\paragraph{Remark.}
Although $D^2 \Phi_L^{\mathrm{tot}} = \sum_{\ell} D^2 \Phi_\ell$ is additive,
the Gaussian curvature depends on determinants of these matrices.
Since the determinant is nonlinear, curvature does not decompose
additively across layers in general. Instead, cross-layer interaction
terms arise, and curvature reflects the combined second-order structure
of all layers.

\newpage

\begin{corollary}[Aligned scalar regime]
Under the assumptions of the theorem, suppose there exist:

\begin{itemize}
\item a function $\psi \in C^2(U)$,
\item scalars $a_1,\dots,a_L$ with $a_\ell \ge a_0 > 0$,
\item a fixed vector $c \in \mathbb{R}^{D-2}$,
\end{itemize}

such that
\[
\Phi_\ell(u,v) = a_\ell \, \psi(u,v)\, c.
\]

Assume $\nabla \psi(0)=0$, and define $A_L := \sum_{\ell=1}^L a_\ell$.
Then
\[
K_{M_L}\bigl(F_L(0)\bigr)
=
A_L^2 \,\|c\|^2 \, \det D^2 \psi(0),
\]
and in particular,
\[
|K_{M_L}(F_L(0))| \ge a_0^2 L^2 \,\|c\|^2\, |\det D^2 \psi(0)|.
\]
\end{corollary}

In this aligned setting, all second-order contributions lie in a common direction,
eliminating cross-layer interaction terms and yielding coherent quadratic scaling. {[Proof. See Appendix~A.1.10]}

\paragraph{From existence to robustness.}
The preceding results establish the existence of positively curved realizations induced by multiplicative gating, their realization within standard attention architectures, and their amplification under composition across layers. However, these results do not rule out the possibility that such curvature arises only from finely tuned parameter choices.

We strengthen this result by showing that positive curvature is stable under perturbations of the gating parameters. In particular, the set of parameters inducing strictly positively curved statistical manifolds contains a non-empty open subset of parameter space. This demonstrates that the curvature gap is a robust and locally generic phenomenon, rather than an artifact of a specific construction.

The following theorem formalizes this stability and genericity property.

\begin{theorem}[Curvature gap and local robustness under multiplicative gating]
Let $D \ge 3$ and $m \ge D$. There exist
\begin{itemize}
    \item a nonempty connected open set $U \subset \mathbb{R}^2$,
    \item an affine map $Y:U \to \mathbb{R}^D$, $Y(\phi)=a+B\phi$, with $a\in\mathbb{R}^D$, $B\in\mathbb{R}^{D\times 2}$, and $\operatorname{rank}(B)=2$,
    \item a smooth map $X^*:U\to\mathbb{R}^m$,
    \item and the logistic sigmoid $\sigma(t)=\frac{1}{1+e^{-t}}$, applied componentwise,
\end{itemize}
such that for each $W_\theta\in\mathbb{R}^{m\times D}$, the map
\[
\mu_{W_\theta}(\phi)
:=
Y(\phi)\odot \sigma\!\big(X^*(\phi)W_\theta\big)
\]
defines a Gaussian location family with identity covariance,
\[
\mathcal{F}(W_\theta)
:=
\bigl\{\mathcal{N}(\mu_{W_\theta}(\phi),I_D):\phi\in U\bigr\}.
\]

On any open subset $V \subset U$ on which $D\mu_{W_\theta}(\phi)$ has rank $2$, this family induces a two-dimensional statistical manifold whose Fisher--Rao metric reduces to
\[
g_{ij}(\phi;W_\theta)
=
\left\langle \partial_i \mu_{W_\theta}(\phi), \partial_j \mu_{W_\theta}(\phi) \right\rangle.
\]

Let
\[
R
:=
\{(\phi,W_\theta)\in U\times \mathbb{R}^{m\times D} : \det g(\phi;W_\theta) > 0\}.
\]

Then:
\begin{enumerate}
    \item[(i)] (\textbf{Flatness of the ungated model})
    If the gate is identically $1$, then $\mu_{W_\theta}(\phi)=Y(\phi)$, the induced metric is constant, and the Gaussian curvature vanishes identically on $U$.

    \item[(ii)] (\textbf{Positive curvature realization})
    There exists $W_\theta^* \in \mathbb{R}^{m\times D}$ such that
    \[
    K(\phi;W_\theta^*) \equiv 1
    \qquad \forall \phi \in U.
    \]
    Thus multiplicative gating strictly enlarges the class of realizable intrinsic geometries, enabling non-zero curvature that affine (ungated) models cannot represent.

    \item[(iii)] (\textbf{Local robustness})
    There exist a nonempty compact set $K \subset U$, a constant $c>0$, and an open neighborhood $\mathcal{O}$ of $W_\theta^*$ such that

    \[
\begin{aligned}
(\phi, W_\theta) &\in R, \\
K(\phi; W_\theta) &\ge \frac{c}{2} > 0 
\quad \forall \phi \in K,\; W_\theta \in \mathcal{O}.
\end{aligned}
\]

\end{enumerate}
\end{theorem}

\begin{lemma}[Regular locus and continuity of curvature]
\label{lem:regular}
Let $R := \{(\phi,W_\theta):\det g(\phi;W_\theta)>0\}$. Then $R$ is open, curvature is well-defined on $R$, and $(\phi,W_\theta)\mapsto K(\phi;W_\theta)$ is continuous on $R$.
\end{lemma}

\begin{lemma}[Uniform robustness on compact sets]
\label{lem:robust}
Let $K \subset U$ be compact. Suppose
\[
(\phi,W_\theta^*)\in R
\quad\text{and}\quad
K(\phi;W_\theta^*) \ge c > 0
\quad \forall \phi \in K.
\]
Then there exists an open neighborhood $\mathcal{O}$ of $W_\theta^*$ such that

\[
\begin{aligned}
(\phi,W_\theta) &\in R, \\
K(\phi;W_\theta) &\ge \frac{c}{2} > 0
\quad \forall \phi \in K,\; W_\theta\in\mathcal{O}.
\end{aligned}
\]
\end{lemma}

\paragraph{From robustness to genericity.}
Theorem~3.17 establishes that strictly positive curvature arises on a non-empty open subset of the gating parameter space, showing that the curvature gap is locally robust. However, this result does not address whether non-flat geometry is typical among admissible gating functions. We strengthen this conclusion by showing that, at the level of local curvature, non-flatness is in fact generic in the sense of Baire category. Within the space of smooth gating functions that induce an immersion at a point, those yielding nonzero curvature form an open dense subset.

\begin{theorem}[Pointwise generic non-flatness under multiplicative gating]
Let $U \subset \mathbb{R}^d$ be open with $d \ge 2$. Let
\[
\mu_g(\phi) = Y(\phi) \odot g(\phi), \quad g \in C^2(U,\mathbb{R}^D),
\]
where
\[
Y(\phi) = a + B\phi, \quad a \in \mathbb{R}^D,\; B \in \mathbb{R}^{D \times d},\; \operatorname{rank}(B)=d.
\]

Fix $\phi_0 \in U$, and assume that at least $d+1$ coordinates of $Y(\phi_0)$ are nonzero. Define
\[
\mathcal I_{\phi_0} := \{ g \in C^2(U,\mathbb{R}^D) : \operatorname{rank} D\mu_g(\phi_0) = d \},
\]
and
\[
\mathcal N_{\phi_0} := \{ g \in \mathcal I_{\phi_0} : R_g(\phi_0) \neq 0 \}.
\]

Then $\mathcal N_{\phi_0}$ is open and dense in $\mathcal I_{\phi_0}$ in the $C^2$ topology.
\end{theorem}

\paragraph{Remark on assumptions.}
The immersion condition ensures that the induced metric is nondegenerate at $\phi_0$, while the nonzero coordinate assumption guarantees the existence of a normal direction accessible through multiplicative perturbations.

\paragraph{Interpretation.}
Theorem~3.20 shows that, in the abstract multiplicative model $\mu_g(\phi) = Y(\phi) \odot g(\phi)$, non-flatness at a fixed immersed point is generic in the sense of open density in the $C^2$ topology. Thus, once the affine restriction of ungated attention is removed, flatness is not structurally enforced, and local intrinsic curvature can arise under arbitrarily small perturbations.

\paragraph{Relation to Theorem~3.17.}
Theorem~3.17 establishes that strictly positive curvature arises on a nonempty open subset of a concrete parameterized gating family. In contrast, Theorem~3.20 shows that non-flatness at a point is open and dense in the broader function-space model. Together, these results demonstrate both the robustness of explicit curved constructions and the prevalence of local non-flat geometry within the unrestricted multiplicative framework.

\section{Experiments}

Our theoretical analysis shows that ungated attention induces intrinsically flat statistical manifolds under idealized assumptions. In practice, attention outputs depend nonlinearly on the input, so even ungated models exhibit small but nonzero curvature when measured with respect to input perturbations. We therefore interpret curvature comparatively, using ungated attention as a low-curvature baseline and multiplicative gating as a mechanism that induces higher curvature. This allows us to test whether the geometric expressivity gap predicted by theory appears in learned models.

We evaluate this on a controlled synthetic sequence classification task designed to require curved representations. We compare ungated and gated attention variants, measure representation curvature using finite-difference proxies, and analyze its relationship with performance under different metric conditions.

\subsection{Experimental Setup}

We evaluate attention geometry on a synthetic sequence classification task with a curved decision rule. Each example is generated by sampling a latent center $c \in [-2,2]^2$ and forming a sequence of length $8$ as noisy observations,
\[
x_i = c + \epsilon_i, \quad \epsilon_i \sim \mathcal{N}(0, \sigma^2 I).
\]
The label depends only on $c$. Let $r = \lVert c \rVert$ and $\theta = \mathrm{atan2}(c_2, c_1)$, and define
\[
s(c) = \sin(2.5\,\theta) + 0.6\,(r - 1.2).
\]
We set $y = 1$ if $s(c) > 0$ and $y = 0$ otherwise.

We use a minimal attention-based model with a single attention block and a classifier head. Inputs in $\mathbb{R}^2$ are projected to $64$ dimensions, processed by scaled dot-product attention, and mean pooled before a two-layer MLP.

We consider variants of the attention output, including ungated attention, a pointwise SiLU nonlinearity, and multiplicative gating. In gated models, we vary a gate strength parameter $\alpha$ that controls the degree of multiplicative modulation,
\[
Y' = Y \odot \left(1 + \alpha \big(\sigma(WY) - 1\big)\right).
\]

When $\alpha = 1$, this reduces to the multiplicative gating form analyzed in the theoretical section.

A residual connection and layer normalization follow the attention block.

Models are trained with AdamW and evaluated over multiple seeds. We measure curvature using a finite-difference curvature proxy
\[
\kappa(x) = \left\lVert \frac{f(x+\epsilon v) - 2f(x) + f(x-\epsilon v)}{\epsilon^2} \right\rVert,
\]
averaged over random directions $v$. This quantity is a finite-difference proxy for the local nonlinearity of the representation map $f$, measuring the magnitude of second-order variation along input directions. It is not an invariant notion of Riemannian curvature and does not directly estimate sectional curvature. Rather, it provides a practical surrogate for comparing the relative geometric complexity of representations across model variants. In particular, higher values of this proxy indicate deviations from locally affine behavior, which is consistent with the presence of nonzero intrinsic curvature predicted by the theory. Our primary metric is curvature of the attention output mean, and we also measure output curvature using a Fisher–Rao square-root embedding. Robustness is evaluated under diagonal precision matrices with condition numbers $2,4,8,12,$ and $20$.
\subsection{Results}

\begin{figure}[H]
    \centering
    \includegraphics[width=0.75\linewidth]{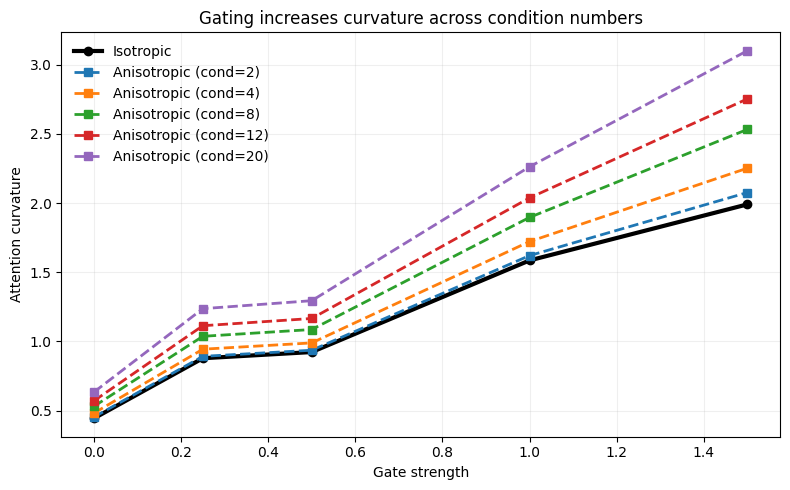}
    \caption{\textbf{Gating increases representation curvature.} Isotropic curvature is invariant across condition numbers, while anisotropic curvature varies with conditioning. Higher gate strength consistently increases curvature.}
    \label{fig:gate_vs_curvature}
\end{figure}

Figure~\ref{fig:gate_vs_curvature} shows curvature as a function of gate strength across condition numbers. Increasing the strength of multiplicative gating leads to a consistent increase in the curvature of the attention representation. The isotropic curvature curves coincide across all condition numbers, indicating that the geometric structure is intrinsic to the representation mapping rather than determined by the metric. In contrast, anisotropic curvature increases with the condition number but preserves the same monotonic dependence on gate strength. These results show that gating controls the curvature of the representation, while the metric primarily affects its scale.

\begin{figure}[H]
    \centering
    \includegraphics[width=1\linewidth]{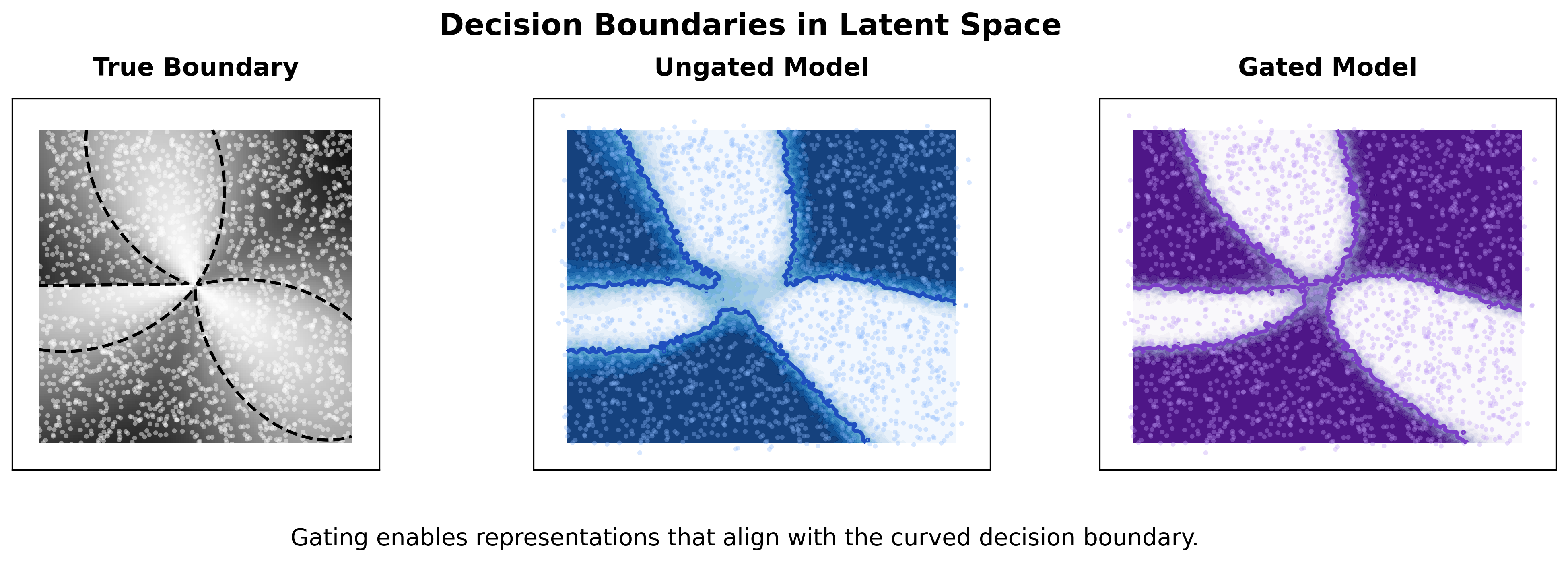}
    \caption{Decision boundaries in latent space on the synthetic curved classification task.
The ungated model fails to align with the nonlinear boundary, while the gated model
better captures the curved structure, demonstrating improved geometric expressivity.}
    \label{fig:decision_boundaries}
\end{figure}

Figure~\ref{fig:decision_boundaries} shows the decision boundaries learned by ungated and gated models on the synthetic curved classification task. The ground-truth boundary exhibits a non-separable radial–angular nonlinear structure, arising from coupled dependence on both radius and angle in polar coordinates. The ungated model fails to capture this geometry, producing approximately piecewise linear regions that deviate from the true boundary in regions requiring higher curvature. In contrast, the gated model yields decision boundaries that more closely align with the underlying nonlinear structure.

\begin{figure}[H]
    \centering
    \includegraphics[width=0.75\linewidth]{imaaaaaaage.png}
    \caption{\textbf{Curvature correlates with task performance.} Test accuracy versus isotropic attention curvature for different gate strengths. Higher curvature leads to improved accuracy, with mild saturation at larger values.}
    \label{fig:curvature_vs_accuracy}
\end{figure}

\paragraph{Curvature correlates with task performance.}
Figure~\ref{fig:curvature_vs_accuracy} plots test accuracy against isotropic representation curvature for different gate strengths. We observe a strong positive correlation between curvature and performance, with a Pearson correlation coefficient of $0.7886$. Accuracy increases with curvature and exhibits mild saturation at higher values. These results indicate that greater geometric expressivity of the representation translates into improved task performance. We report isotropic curvature here to isolate the intrinsic geometry of the representation. Corresponding results under anisotropic metrics are provided in the appendix and exhibit consistent trends.

\paragraph{Linear control task.}
To assess whether the observed gains arise from generic additional nonlinearity, we also evaluate a control task with a linear decision boundary that does not require curved representations. In this setting, gating does not provide a consistent advantage (see Appendix~A.4), indicating that its benefits arise specifically in settings requiring nonlinear geometric structure.

\subsection{Ablation study}

\begin{figure}[H]
    \centering
    \includegraphics[width=0.75\linewidth]{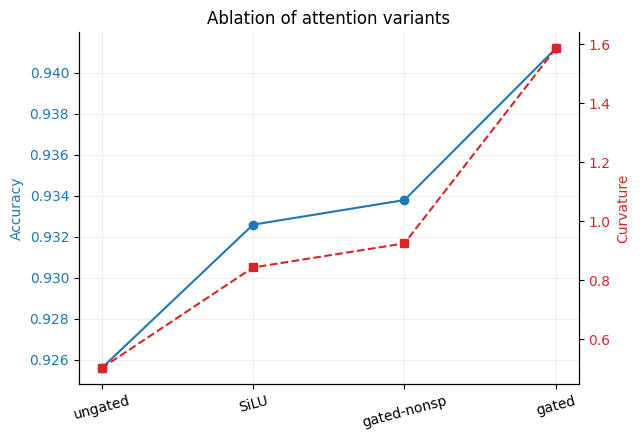}
    \caption{\textbf{Ablation of attention variants.} Test accuracy (left axis) and isotropic curvature (right axis) for different attention variants. Ungated attention yields the lowest curvature and accuracy. Adding a pointwise SiLU nonlinearity increases both modestly, while multiplicative gating produces substantially higher curvature and improved accuracy. This shows that gains in geometric expressivity arise specifically from multiplicative gating rather than generic nonlinear transformations.}
    \label{fig:ablation}
\end{figure}

\paragraph{Effect of multiplicative gating.}
We compare attention variants to isolate the role of multiplicative gating. Let $Y$ denote the attention output. The ungated variant uses $Y$ directly. The SiLU variant applies the pointwise nonlinearity
\[
Y' = \mathrm{SiLU}(Y),
\qquad
\mathrm{SiLU}(x) = x \cdot \sigma(x),
\]
where $\sigma$ denotes the sigmoid function. The gated-sigmoid variant applies multiplicative modulation of the form
\[
Y' = Y \odot \sigma(WY),
\]
and the gated-nonsparse variant uses the same multiplicative structure with a rescaled gate
\[
Y' = Y \odot \left(0.5 + 0.5\,\sigma(WY)\right).
\]
In the ablation study, all gated variants use gate strength $1$ for consistency. Figure~\ref{fig:ablation} shows test accuracy and isotropic curvature across variants. Ungated attention yields the lowest curvature and accuracy, while a SiLU nonlinearity provides only modest improvements. In contrast, multiplicative gating produces substantially higher curvature and improved accuracy, indicating that geometric expressivity arises specifically from gating rather than generic nonlinear transformations. We report isotropic curvature to isolate representation effects, and include anisotropic results in the appendix, which show consistent trends.

These observations are consistent with the geometric interpretation, where gating better captures nonlinear structure.

\subsection{Conclusion}

We study the geometry of attention representations through curvature and show that multiplicative gating strictly enlarges the class of geometries realizable by attention, enabling statistical manifolds with nonzero intrinsic curvature that are unattainable in the ungated setting. Furthermore, we identify a structured regime in which curvature can accumulate under composition, yielding a depth-dependent amplification effect.

Empirically, we find that curvature is significantly higher with multiplicative gating compared to other variants, such as ungated attention and pointwise nonlinear attention. We observe that isotropic curvature is positively correlated with task performance, with mild saturation at higher values.

We also find that curvature is an intrinsic property of the representation mapping. Ablation studies confirm that the observed gains arise specifically from multiplicative gating rather than generic nonlinear transformations.

\bibliographystyle{abbrv}
\bibliography{references}

\newpage
\appendix
\onecolumn
\section{Appendix}
\subsection{Proofs of Main Results}

\subsubsection*{A.1.1 Proof of Theorem~3.1}

\begin{proof}
The attention output takes the form
\[
\mu = \sum_{i=1}^n \alpha_i U_i, \quad \alpha \in \Delta_{n-1},
\]
so the set of outputs lies in the convex hull of $\{U_1, \dots, U_n\}$.

On the relative interior, we can eliminate one coordinate using the constraint $\sum_i \alpha_i = 1$. Writing
\[
\alpha_n = 1 - \sum_{i=1}^{n-1} \alpha_i,
\]
we obtain
\[
\mu = \sum_{i=1}^{n-1} \alpha_i U_i + \left(1 - \sum_{i=1}^{n-1} \alpha_i\right) U_n
= U_n + \sum_{i=1}^{n-1} \alpha_i (U_i - U_n).
\]
Let
\[
k = \dim \operatorname{aff}\{U_1,\dots,U_n\}.
\]
On the relative interior of the convex hull, choose local affine coordinates
\[
\phi \in \Omega \subset \mathbb{R}^k,
\]
under which the map admits an affine parameterization
\[
\mu_{\mathrm{ung}}(\phi) = a + B\phi,
\]
where \(a \in \mathbb{R}^D\) and \(B \in \mathbb{R}^{D \times k}\) has full column rank \(k\).
Its partial derivatives are constant:
\[
\partial_i \mu_{\mathrm{ung}}(\phi) = B_i \in \mathbb{R}^D,
\]
where $B_i$ denotes the $i$-th column of $B$.

Equipping the Gaussian family $\mathcal{N}(\mu, I_D)$ with the Fisher--Rao metric, the induced Riemannian metric on the parameter domain $\Omega$ is
\[
g_{ij}^{\mathrm{ung}}(\phi) = \langle \partial_i \mu_{\mathrm{ung}}(\phi), \partial_j \mu_{\mathrm{ung}}(\phi) \rangle_{\mathbb{R}^D}
= \langle B_i, B_j \rangle_{\mathbb{R}^D} =: C_{ij}.
\]
Since \(B \in \mathbb{R}^{D \times k}\) has full column rank, the matrix
\[
C = B^\top B
\]
is symmetric positive definite. Hence, the metric tensor is independent of $\phi$.

For a constant metric tensor in global coordinates, the Christoffel symbols of the Levi--Civita connection vanish identically:
\[
\Gamma^k_{ij} = \tfrac{1}{2} g^{k\ell} \big( \partial_i g_{j\ell} + \partial_j g_{i\ell} - \partial_\ell g_{ij} \big) = 0.
\]
Consequently, the Riemann curvature tensor,
\[
R^k_{\;\ell ij} = \partial_i \Gamma^k_{\ell j} - \partial_j \Gamma^k_{\ell i}
+ \Gamma^k_{im} \Gamma^m_{\ell j} - \Gamma^k_{jm} \Gamma^m_{\ell i},
\]
vanishes identically. Therefore, all sectional curvatures are zero, and the statistical manifold $\mathcal{M}_{\mathrm{ung}}$ is intrinsically flat.
\end{proof}

\subsubsection*{A.1.2 Proof of Lemma~3.4}

\begin{proof}
Differentiation gives the stated formula. Define
\[
T: C^2(U,\mathbb{R}^D)\to C^0(U,\mathbb{R}^{D\times d\times d}),
\qquad
T(g)=D^2(Y\odot g).
\]
Then $T$ is continuous and linear, and $\mathcal A=\ker T$ since $\mu$ is affine if and only if $D^2\mu\equiv 0$ on connected $U$. Thus $\mathcal A$ is a closed linear subspace.

To see $\mathcal A\neq C^2(U,\mathbb{R}^D)$, note that since $\mathrm{rank}(B)=d\ge 1$, we have $B\neq 0$, hence $Y(\phi)=a+B\phi$ is not identically zero. Therefore there exist $m\in\{1,\dots,D\}$ and $\phi_0\in U$ such that $Y_m(\phi_0)\neq 0$. Take $g=e_m\psi$ with $\psi\in C_c^\infty(U)$ satisfying
\[
\psi(\phi_0)=0,\qquad \partial_i\psi(\phi_0)=0,\qquad \partial_{11}\psi(\phi_0)=1.
\]
Then
\[
(Tg)_{11,m}(\phi_0)
=
Y_m(\phi_0)\,\partial_{11}\psi(\phi_0)
=
Y_m(\phi_0)\neq 0,
\]
so $g\notin\ker T$. Hence $\mathcal A$ is a proper closed linear subspace, and therefore nowhere dense and meagre.
\end{proof}

\subsubsection*{A.1.3 Proof of Theorem~3.5}

\begin{proof}
We construct the example explicitly.

Let
\[
Y(\phi)
=
\begin{pmatrix}
2+\phi_1\\
2+\phi_2\\
2
\end{pmatrix},
\qquad
\phi=(\phi_1,\phi_2)\in \mathbb{R}^2.
\]
This is an affine map of the form
\[
Y(\phi)=a+B\phi,
\qquad
a=
\begin{pmatrix}
2\\2\\2
\end{pmatrix},
\qquad
B=
\begin{pmatrix}
1&0\\
0&1\\
0&0
\end{pmatrix},
\]
with $\operatorname{rank}(B)=2$.

Since the decoder is the Gaussian location family
\[
p(y\mid \mu)=\mathcal N(y;\mu,I_3),
\]
the Fisher--Rao metric on mean-parameter space coincides with the Euclidean metric on $\mathbb{R}^3$. Therefore the statistical manifold induced by the ungated map
\[
\mu_{\mathrm{ung}}(\phi):=Y(\phi)
\]
is the image of an affine embedding of a two-dimensional domain into $\mathbb{R}^3$, hence is intrinsically flat. Thus
\[
K_{\mathrm{ung}}\equiv 0.
\]

Choose
\[
\phi_0=(0,0),
\qquad
Y(\phi_0)=
\begin{pmatrix}
2\\2\\2
\end{pmatrix}\neq 0.
\]
Since $Y$ is continuous, there exists a sufficiently small connected open neighborhood $U\subset \mathbb{R}^2$ of $\phi_0$ such that $\|Y(\phi)\|$ is bounded away from zero on $U$. Hence the map
\[
\phi\longmapsto \frac{1}{\|Y(\phi)\|}
\]
is smooth on $U$. After shrinking $U$ further if necessary, we may also assume
\[
0<\frac{1}{\|Y(\phi)\|}<1
\qquad
\text{for all }\phi\in U.
\]
This ensures that the inverse of the activation function used below is well-defined on the relevant range.

Now define the normalized map
\[
\mu_{\mathrm{sph}}(\phi):=\frac{Y(\phi)}{\|Y(\phi)\|}.
\]
By construction, $\|\mu_{\mathrm{sph}}(\phi)\|=1$ for all $\phi\in U$, so its image lies in the unit sphere $S^2$.

We claim that, after possibly shrinking $U$ once more, $\mu_{\mathrm{sph}}$ parametrizes a smooth spherical patch. Consider the normalization map
\[
N:\mathbb{R}^3\setminus\{0\}\to S^2,
\qquad
N(y)=\frac{y}{\|y\|}.
\]
Its restriction to the affine plane
\[
\Pi:=\{(y_1,y_2,y_3)\in\mathbb{R}^3:\ y_3=2\}
\]
has differential of rank $2$ at every point of $\Pi$. Since $\Pi$ is two-dimensional, it follows from the inverse function theorem that $N|_\Pi$ is a local diffeomorphism onto its image. Because $Y(U)\subset \Pi$, after shrinking $U$ if necessary the map
\[
\mu_{\mathrm{sph}} = N\circ Y
\]
parametrizes a smooth patch of $S^2$.

We now realize this normalization exactly through multiplicative gating. Let
\[
g(\phi):=\frac{1}{\|Y(\phi)\|}\mathbf{1}
=
\begin{pmatrix}
\|Y(\phi)\|^{-1}\\
\|Y(\phi)\|^{-1}\\
\|Y(\phi)\|^{-1}
\end{pmatrix},
\qquad
\mathbf{1}=
\begin{pmatrix}
1\\1\\1
\end{pmatrix}.
\]
Then
\[
Y(\phi)\odot g(\phi)
=
\frac{Y(\phi)}{\|Y(\phi)\|}
=
\mu_{\mathrm{sph}}(\phi).
\]

It remains to express $g$ in the form $\sigma(X(\phi)W_\theta)$ exactly. Choose a smooth strictly monotone activation
\[
\sigma:\mathbb{R}\to (0,1),
\]
for example the logistic sigmoid
\[
\sigma(t)=\frac{1}{1+e^{-t}}.
\]
Since $\|Y(\phi)\|^{-1}\in (0,1)$ on $U$, the inverse function $\sigma^{-1}$ is well-defined and smooth on the range of $\|Y(\phi)\|^{-1}$.

Define
\[
X(\phi):=\sigma^{-1}\!\left(\frac{1}{\|Y(\phi)\|}\right)\in \mathbb{R},
\qquad
W_\theta=
\begin{pmatrix}
1&1&1
\end{pmatrix}\in \mathbb{R}^{1\times 3}.
\]
Then $\operatorname{rank}(W_\theta)=1$, and
\[
X(\phi)W_\theta
=
\begin{pmatrix}
X(\phi)&X(\phi)&X(\phi)
\end{pmatrix}.
\]
Applying $\sigma$ componentwise gives
\[
\sigma\!\bigl(X(\phi)W_\theta\bigr)
=
\begin{pmatrix}
\sigma(X(\phi))\\
\sigma(X(\phi))\\
\sigma(X(\phi))
\end{pmatrix}
=
\begin{pmatrix}
\|Y(\phi)\|^{-1}\\
\|Y(\phi)\|^{-1}\\
\|Y(\phi)\|^{-1}
\end{pmatrix}
=
g(\phi).
\]
Therefore
\[
\mu_{\mathrm{gat}}(\phi)
:=
Y(\phi)\odot \sigma\!\bigl(X(\phi)W_\theta\bigr)
=
Y(\phi)\odot g(\phi)
=
\frac{Y(\phi)}{\|Y(\phi)\|}
=
\mu_{\mathrm{sph}}(\phi).
\]

Thus the gated map parametrizes a smooth patch of the unit sphere. Since the sphere $S^2$ with its induced Euclidean metric has constant Gaussian curvature equal to $1$, and since in the Gaussian location family the Fisher--Rao metric is exactly the Euclidean pullback metric on the mean manifold, the induced statistical manifold satisfies
\[
K_{\mathrm{gat}}(\phi)\equiv 1>0
\qquad
\text{for all }\phi\in U.
\]

Hence the ungated statistical manifold is flat, while the gated statistical manifold has strictly positive curvature. This proves the theorem.
\end{proof}

\subsubsection*{A.1.4 Proof of Theorem~3.7}

\begin{proof}
The proof gives an explicit witness inside a standard content-aware transformer block.

Let
\[
U:=\left(0,\frac{\pi}{4}\right)\times\left(0,\frac{\pi}{4}\right)\subset\mathbb R^2,
\qquad
\phi=(\phi_1,\phi_2).
\]
Define
\[
s(\phi_1,\phi_2)=
\begin{pmatrix}
\cos\phi_1\cos\phi_2\\
\cos\phi_1\sin\phi_2\\
\sin\phi_1
\end{pmatrix}.
\]
Since \(\phi_1,\phi_2\in(0,\pi/4)\), all three coordinates of \(s(\phi)\) are strictly positive, so
\[
s(\phi)\in (0,1)^3
\qquad
\text{for all }\phi\in U.
\]
This is the standard local parametrization of a patch of the unit sphere \(S^2\subset\mathbb R^3\), and its induced Gaussian curvature is identically \(1\).

\medskip
\noindent
\textbf{Step 1: Enforce affine structure in the ungated output.}
Set
\[
W_Q=0,\qquad W_K=0.
\]

Note that content-dependent attention mechanisms strictly contain this uniform-weight case as a special instance. Therefore, any geometric expressivity achievable under this construction is realizable within the full content-aware attention class.

Then \(Q(X)=0\) and \(K(X)=0\) for every input \(X\), so all attention scores are equal and the softmax produces uniform weights
\[
\alpha_{ji}(X)=\frac1n
\qquad
\text{for all }X,\ i=1,\dots,n.
\]

Choose \(W_V\) and \(W_O\) so that the composed linear map
\[
L:=W_VW_O
\]
has rank \(3\). Fix a vector \(a\in\mathbb R^3\) and a matrix \(B\in\mathbb R^{3\times 2}\) of rank \(2\), chosen so that
\[
0 < \frac{s_k(\phi)}{(a+B\phi)_k} < 1
\qquad
\text{for all }\phi\in U,\ k=1,2,3.
\]
Such a choice is possible because \(s(\phi)\) is bounded on the relatively compact set \(\overline U\) (closure of U), so choosing \(a\) sufficiently large componentwise ensures
\[
(a+B\phi)_k > s_k(\phi) >0
\qquad
\text{for all }\phi\in U,\ k=1,2,3.
\]

Now define
\[
U_1(X(\phi)):=n(a+B\phi),
\qquad
U_i(X(\phi)):=0 \quad \text{for } i=2,\dots,n.
\]
Because \(L\) has rank \(3\), the input token embeddings \(X(\phi)\) can be chosen smoothly so that these value vectors are realized exactly. With these choices,
\[
Y_j(X(\phi))
=
\sum_{i=1}^n \alpha_{ji}(X(\phi))U_i(X(\phi))
=
\frac1n\,U_1(X(\phi))
=
a+B\phi.
\]
Hence the ungated map \(\phi\mapsto Y_j(X(\phi))\) is affine. Since the decoder is the Gaussian location family with covariance \(I_3\), the Fisher--Rao metric on mean space is Euclidean, and therefore the induced manifold
\[
\mathcal M_{\mathrm{ung}}
=
\{\mathcal N(Y_j(X(\phi)),I_3):\phi\in U\}
\]
is isometric to an open subset of Euclidean space. Thus
\[
K_{\mathrm{ung}}(\phi)=0
\qquad
\text{for all }\phi\in U.
\]

\medskip
\noindent
\textbf{Step 2: Construct an exact multiplicative gate realizing a sphere.}
Define the componentwise ratio
\[
\tilde z(\phi):=\frac{s(\phi)}{a+B\phi},
\]
where the division is taken coordinatewise. By the choice of \(a\) and \(B\),
\[
\tilde z(\phi)\in (0,1)^3
\qquad
\text{for all }\phi\in U.
\]

Let \(\sigma^{-1}:(0,1)\to\mathbb R\) denote the inverse of the scalar nonlinearity, applied coordinatewise, and define
\[
t(\phi):=\sigma^{-1}(\tilde z(\phi))\in\mathbb R^3.
\]
Then \(t\) is smooth and satisfies
\[
\sigma(t(\phi))=\tilde z(\phi)
\qquad
\text{for all }\phi\in U.
\]

Choose any matrix
\[
W_\theta\in\mathbb R^{m\times 3}
\qquad
\text{with }
\operatorname{rank}(W_\theta)=3.
\]
Since $W_\theta \in \mathbb{R}^{m\times 3}$ has full column rank, it admits a left inverse $W_\theta^\dagger \in \mathbb{R}^{3\times m}$ such that
\[
W_\theta^\dagger W_\theta = I_3.
\]
Let $t(\phi) \in \mathbb{R}^3$ and view $t(\phi)^\top \in \mathbb{R}^{1\times 3}$. Define
\[
X_g(\phi) := t(\phi)^\top W_\theta^\dagger \in \mathbb{R}^{1\times m}.
\]
Then
\[
X_g(\phi) W_\theta = t(\phi)^\top W_\theta^\dagger W_\theta = t(\phi)^\top,
\]
and therefore
\[
\sigma(X_g(\phi)W_\theta) = \sigma(t(\phi)) = \tilde z(\phi)
\]
holds exactly. Consequently,
\[
Y'_j(X(\phi)) = Y_j(X(\phi)) \odot \sigma(X_g(\phi)W_\theta)
= (a + B\phi) \odot \tilde z(\phi) = s(\phi).
\]
Thus the gated output coincides exactly with the spherical parametrization $s(\phi)$.

\medskip
\noindent
\textbf{Step 3: Deduce curvature.}
Since \(s(\phi)\) parametrizes a smooth patch of the unit sphere \(S^2\), the induced Euclidean metric on its image is the spherical metric, whose Gaussian curvature is identically \(1\). Because the Fisher--Rao metric for the Gaussian location family coincides with the Euclidean pullback metric on the mean manifold, the induced gated statistical manifold
\[
\mathcal M_{\mathrm{gat}}
=
\{\mathcal N(Y'_j(X(\phi)),I_3):\phi\in U\}
\]
is isometric to a spherical patch and satisfies
\[
K_{\mathrm{gat}}(\phi)=1
\qquad
\text{for all }\phi\in U.
\]

Combining the flatness of \(\mathcal M_{\mathrm{ung}}\) with the positive curvature of \(\mathcal M_{\mathrm{gat}}\) yields the claimed curvature gap inside standard content-aware attention. The construction above is an explicit witness that multiplicative gating strictly enlarges the class of realizable representation geometries in this architectural family.
\end{proof}

\subsubsection*{A.1.5 Proof of Lemma~3.10}

\begin{proof}
By linearity of summation,
\[
\sum_{\ell=1}^L a_\ell \psi(u,v)
=
\Bigl(\sum_{\ell=1}^L a_\ell\Bigr)\psi(u,v)
=
A_L \psi(u,v).
\]
Substituting this identity into the chart expression yields
\[
\chi_{\mathrm{out}} \circ F_L \circ \chi_{\mathrm{in}}^{-1}(u,v)
=
\bigl(u,\ v,\ A_L \psi(u,v),\ 0,\dots,0\bigr).
\]
This proves the claim.
\end{proof}

\subsubsection*{A.1.6 Proof of Lemma~3.11}

\begin{proof}
We construct the model explicitly.

Fix an ambient dimension $D \ge 4$. Define the initial state
\[
h^{(0)}(u,v) = (u, v, 0, 1, 0, \dots, 0) \in \mathbb R^D,
\]
and define the input parametrization by
\[
X(u,v) = h^{(0)}(u,v).
\]

The fourth coordinate is a constant auxiliary feature.

\textbf{Attention mechanism.}
We instantiate each attention block on a single-token input, so the softmax weight is identically equal to $1$. This is a valid special case of the standard attention mechanism.

Choose value and output projections so that the ungated attention output is exactly
\[
Y(u,v) = e_3 = (0,0,1,0,\dots,0).
\]
This is achieved by letting the value/output map read only the constant fourth coordinate and send it to the third coordinate.

\textbf{Gate construction.}
Define
\[
X_g^{(\ell)}(u,v) = \operatorname{logit}(a_\ell \psi(u,v))\, e_3,
\qquad
W_\theta^{(\ell)} = I_D.
\]

Then
\[
X_g^{(\ell)}(u,v) W_\theta^{(\ell)}
=
\operatorname{logit}(a_\ell \psi(u,v))\, e_3.
\]

Applying $\sigma$ componentwise yields
\[
\sigma\!\bigl(X_g^{(\ell)} W_\theta^{(\ell)}\bigr)
=
\bigl(\tfrac12,\tfrac12,a_\ell \psi(u,v),\tfrac12,\dots,\tfrac12\bigr).
\]

\textbf{Gated output.}
Since $Y(u,v)=e_3$, we obtain
\[
Y'(u,v) = e_3 \odot \sigma(\cdot) = a_\ell \psi(u,v)\, e_3.
\]

\textbf{Residual accumulation.}
Define
\[
h^{(\ell)} = h^{(\ell-1)} + a_\ell \psi(u,v)\, e_3.
\]

By induction,
\[
h^{(L)}(u,v)
=
\bigl(u,\ v,\ \sum_{\ell=1}^L a_\ell \psi(u,v),\ 1,\ 0,\dots,0\bigr).
\]

\textbf{Final output map.}
Define a linear projection $P : \mathbb R^D \to \mathbb R^D$ by
\[
P(x_1,x_2,x_3,x_4,x_5,\dots,x_D) = (x_1,x_2,x_3,0,\dots,0).
\]

Then
\[
P\bigl(h^{(L)}(u,v)\bigr)
=
\bigl(u,\ v,\ \sum_{\ell=1}^L a_\ell \psi(u,v),\ 0,\dots,0\bigr),
\]
which is the claimed normal form.
\end{proof}

\subsubsection*{A.1.7 Proof of Lemma~3.12}

\begin{proof}
Because the last $D-3$ coordinates are constant, the intrinsic geometry of $M_L$ is the same as that of the graph surface
\[
X(u,v)=(u,v,f_L(u,v))\subset \mathbb R^3,
\qquad
f_L(u,v):=A_L\psi(u,v).
\]
Hence it suffices to compute the Gaussian curvature of the graph surface $X$.

The tangent vectors are
\[
X_u=(1,0,f_u),
\qquad
X_v=(0,1,f_v).
\]
Thus the coefficients of the first fundamental form are
\[
E=\langle X_u,X_u\rangle = 1+f_u^2,
\qquad
F=\langle X_u,X_v\rangle = f_u f_v,
\qquad
G=\langle X_v,X_v\rangle = 1+f_v^2.
\]
A unit normal vector is
\[
N=\frac{(-f_u,-f_v,1)}{\sqrt{1+f_u^2+f_v^2}}.
\]
The second derivatives are
\[
X_{uu}=(0,0,f_{uu}),
\qquad
X_{uv}=(0,0,f_{uv}),
\qquad
X_{vv}=(0,0,f_{vv}),
\]
so the coefficients of the second fundamental form are
\[
e=\langle X_{uu},N\rangle=\frac{f_{uu}}{\sqrt{1+f_u^2+f_v^2}},
\]
\[
f=\langle X_{uv},N\rangle=\frac{f_{uv}}{\sqrt{1+f_u^2+f_v^2}},
\]
\[
g=\langle X_{vv},N\rangle=\frac{f_{vv}}{\sqrt{1+f_u^2+f_v^2}}.
\]
Therefore,
\[
K=\frac{eg-f^2}{EG-F^2}
=
\frac{f_{uu}f_{vv}-f_{uv}^2}{\bigl(1+f_u^2+f_v^2\bigr)^2}.
\]

Now evaluate at the origin. Since $\nabla \psi(0)=0$,
\[
\nabla f_L(0)=A_L\nabla \psi(0)=0,
\qquad
D^2 f_L(0)=A_L D^2\psi(0).
\]
Hence
\[
K_{M_L}\!\bigl(F_L(0)\bigr)
=
\det D^2 f_L(0).
\]
Using the scaling of the determinant for a $2\times 2$ matrix,
\[
\det D^2 f_L(0)
=
\det\!\bigl(A_L D^2\psi(0)\bigr)
=
A_L^2 \det D^2\psi(0).
\]
Thus
\[
K_{M_L}\!\bigl(F_L(0)\bigr)=A_L^2\,\det D^2\psi(0).
\]
\end{proof}

\subsubsection*{A.1.8 Proof of Theorem~3.13}

\begin{proof}
By the local gated-attention normal form lemma,
\[
\chi_{\mathrm{out}} \circ F_L \circ \chi_{\mathrm{in}}^{-1}(u,v)
=
\bigl(u,\ v,\ A_L\psi(u,v),\ 0,\dots,0\bigr).
\]
Therefore the graph-curvature lemma applies and gives
\[
K_{M_L}\!\bigl(F_L(0)\bigr)=A_L^2\,\det D^2\psi(0).
\]

If $a_\ell \ge a_0>0$ for all $\ell$, then
\[
A_L=\sum_{\ell=1}^L a_\ell \ge a_0L.
\]
Taking absolute values yields
\[
\bigl|K_{M_L}\!\bigl(F_L(0)\bigr)\bigr|
=
A_L^2\,\bigl|\det D^2\psi(0)\bigr|
\ge
a_0^2\,\bigl|\det D^2\psi(0)\bigr|\,L^2.
\]
If furthermore $\det D^2\psi(0)>0$, then the absolute values may be dropped and we obtain
\[
K_{M_L}\!\bigl(F_L(0)\bigr)
\ge
a_0^2\,\det D^2\psi(0)\,L^2.
\]
\end{proof}

\subsubsection*{A.1.9 Proof of Theorem~3.15}

\begin{proof}
Write
\[
\Phi_L^{\mathrm{tot}} = (f^1,\dots,f^{D-2}),
\]
so that
\[
F_L(u,v) = (u,v,f^1(u,v),\dots,f^{D-2}(u,v)).
\]

First derivatives are
\[
F_u = (1,0,f_u^1,\dots,f_u^{D-2}), \quad
F_v = (0,1,f_v^1,\dots,f_v^{D-2}).
\]
Since $D\Phi_L^{\mathrm{tot}}(0)=0$, we have $f_u^\alpha(0)=f_v^\alpha(0)=0$,
so
\[
F_u(0)=(1,0,0,\dots,0), \quad F_v(0)=(0,1,0,\dots,0).
\]
Thus the induced metric satisfies $g_{ij}(0)=\delta_{ij}$.

The normal space at $F_L(0)$ is spanned by
\[
n_\alpha = e_{2+\alpha}, \quad \alpha=1,\dots,D-2.
\]

Second derivatives are
\[
F_{uu}=(0,0,f_{uu}^1,\dots,f_{uu}^{D-2}), \quad
F_{uv}=(0,0,f_{uv}^1,\dots,f_{uv}^{D-2}), \quad
F_{vv}=(0,0,f_{vv}^1,\dots,f_{vv}^{D-2}).
\]

Thus the second fundamental form in direction $n_\alpha$ is
\[
II^{(\alpha)} =
\begin{pmatrix}
f_{uu}^\alpha(0) & f_{uv}^\alpha(0) \\
f_{uv}^\alpha(0) & f_{vv}^\alpha(0)
\end{pmatrix}
= D^2 f^\alpha(0).
\]

At the base point, the Gaussian curvature is given by the Gauss equation
\[
K = \sum_{\alpha=1}^{D-2}
\left(h_{11}^\alpha h_{22}^\alpha - (h_{12}^\alpha)^2\right)
= \sum_{\alpha=1}^{D-2} \det\bigl(II^{(\alpha)}\bigr),
\]
where $h_{ij}^\alpha = \langle F_{ij}, n_\alpha \rangle$.

Substituting $II^{(\alpha)} = D^2 f^\alpha(0)$ gives
\[
K_{M_L}\bigl(F_L(0)\bigr)
=
\sum_{\alpha=1}^{D-2}
\det\!\Bigl(D^2(\Phi_L^{\mathrm{tot}})^\alpha(0)\Bigr).
\]

Linearity of the Hessian yields
\[
D^2(\Phi_L^{\mathrm{tot}})^\alpha(0)
=
\sum_{\ell=1}^L D^2 \Phi_\ell^\alpha(0),
\]
which gives the stated formula.
\end{proof}

\subsubsection*{A.1.10 Proof of corollary~3.16}

\begin{proof}
We have
\[
\Phi_L^{\mathrm{tot}}(u,v)
=
\sum_{\ell=1}^L a_\ell \psi(u,v) c
=
A_L \psi(u,v) c.
\]
Thus
\[
D^2(\Phi_L^{\mathrm{tot}})^\alpha(0)
=
A_L c_\alpha D^2 \psi(0).
\]

Using $\det(\lambda H)=\lambda^2 \det(H)$ for $2\times2$ matrices,
\[
\det\bigl(D^2(\Phi_L^{\mathrm{tot}})^\alpha(0)\bigr)
=
A_L^2 c_\alpha^2 \det D^2 \psi(0).
\]

Summing over $\alpha$ gives
\[
K_{M_L}\bigl(F_L(0)\bigr)
=
A_L^2 \|c\|^2 \det D^2 \psi(0).
\]

Since $a_\ell \ge a_0 > 0$, we have $A_L \ge a_0 L$, hence
\[
A_L^2 \ge a_0^2 L^2,
\]
which gives the bound.
\end{proof}

\subsubsection*{A.1.11 Proof of Lemma~3.18}

\begin{proof}
The map $(\phi,W_\theta)\mapsto \mu_{W_\theta}(\phi)$ is smooth, hence $g_{ij}(\phi;W_\theta)$ depends continuously on $(\phi,W_\theta)$. Therefore $\det g$ is continuous and $R$ is open.

Since $g = (D\mu_{W_\theta})^\top D\mu_{W_\theta}$, positivity implies that $D\mu_{W_\theta}$ has rank $2$, so $\mu_{W_\theta}$ is an immersion and curvature is well-defined.

Gaussian curvature depends smoothly on the metric and its derivatives, hence is continuous on $R$.
\end{proof}

\subsubsection*{A.1.12 Proof of Lemma~3.19}

\begin{proof}
Since $(\phi,W_\theta)\mapsto K(\phi;W_\theta)$ is continuous on $R$ (Lemma~\ref{lem:regular}), and
\[
K(\phi;W_\theta^*) \ge c > 0
\quad \forall \phi \in K,
\]
it follows that
\[
\inf_{\phi \in K} K(\phi;W_\theta^*) \ge c.
\]

By continuity in both variables, $K(\phi;W_\theta)$ is uniformly continuous on a neighborhood of the compact set $K \times \{W_\theta^*\}$. Hence there exists an open neighborhood $\mathcal{O}$ of $W_\theta^*$ such that
\[
|K(\phi;W_\theta) - K(\phi;W_\theta^*)| < \frac{c}{2}
\quad \forall \phi \in K,\; W_\theta \in \mathcal{O}.
\]

Therefore
\[
K(\phi;W_\theta) \ge \frac{c}{2} > 0.
\]

Since $R$ is open and contains $(\phi,W_\theta^*)$, possibly shrinking $\mathcal{O}$ ensures $(\phi,W_\theta)\in R$ for all $\phi \in K$, completing the proof.
\end{proof}

\subsubsection*{A.1.13 Proof of Theorem~3.17}

\begin{proof}
For each $W_\theta$, the map $\mu_{W_\theta}:U\to\mathbb{R}^D$ is smooth since $Y$, $X^*$, and $\sigma$ are smooth. Hence
\[
\phi \mapsto \mathcal{N}(\mu_{W_\theta}(\phi),I_D)
\]
defines a smooth Gaussian location family. On any open set where $D\mu_{W_\theta}$ has rank $2$, the map is an immersion. The Fisher--Rao metric reduces to the Euclidean pullback,
\[
g_{ij}(\phi;W_\theta)
=
\langle \partial_i \mu_{W_\theta}(\phi), \partial_j \mu_{W_\theta}(\phi)\rangle.
\]

\paragraph{(i) Flatness.}
If the gate is identically $1$, then
\[
\mu_{W_\theta}(\phi)=Y(\phi)=a+B\phi.
\]
Thus $\partial_i \mu_{W_\theta}=Be_i$ is constant, so the metric is constant. Hence all Christoffel symbols vanish and the Gaussian curvature is identically zero.

\paragraph{(ii) Positive curvature realization.}
Let $s:U\to S^2\subset \mathbb R^3$ be a smooth spherical patch. Define
\[
\tilde s(\phi)=(s_1(\phi),s_2(\phi),s_3(\phi),c_4,\dots,c_D),
\]
with constants $c_4,\dots,c_D>0$. Since the added coordinates are constant, the first fundamental form is unchanged, so $\tilde s$ has curvature identically $1$.

After possibly shrinking $U$, we may assume $\tilde s$ is bounded. Choose the affine map $Y(\phi)=a+B\phi$ so that
\[
0<\tilde s_j(\phi)<Y_j(\phi)\qquad \forall \phi\in U,\; j=1,\dots,D.
\]
Define
\[
z(\phi)=\tilde s(\phi)\oslash Y(\phi)\in(0,1)^D,
\qquad
\ell(\phi)=\sigma^{-1}(z(\phi)).
\]
Set
\[
X^\ast(\phi)=(\ell(\phi),0,\dots,0),
\qquad
W_\theta^\ast=\begin{bmatrix}I_D\\0\end{bmatrix}.
\]
Then $X^\ast(\phi)W_\theta^\ast=\ell(\phi)$, hence
\[
\mu_{W_\theta^\ast}(\phi)
=Y(\phi)\odot \sigma(\ell(\phi))
=\tilde s(\phi).
\]
Thus $K(\phi;W_\theta^\ast)\equiv 1$.

\paragraph{(iii) Local robustness.}
Fix a nonempty compact set $K \subset U$. Since $K(\phi;W_\theta^*)=1$ for all $\phi\in K$, the result follows from Lemma~\ref{lem:robust}.
\end{proof}

\subsubsection*{A.1.14 Proof of theorem~3.20}

\begin{proof}
\textbf{Openness.}
Let $g \in \mathcal N_{\phi_0}$. Since $g \in \mathcal I_{\phi_0}$, the induced metric is nondegenerate at $\phi_0$. On $\mathcal I_{\phi_0}$, the curvature tensor at $\phi_0$ depends smoothly on the $2$-jet of $\mu_g$, hence continuously on $g$ in the $C^2$ topology. Therefore the condition $R_g(\phi_0) \neq 0$ is stable under sufficiently small perturbations, so $\mathcal N_{\phi_0}$ is open.

\textbf{Density.}
Fix $g \in \mathcal I_{\phi_0}$ and set $F := \mu_g$. Let
\[
T := \operatorname{Im} DF(\phi_0).
\]

Let
\[
S := \{ j : Y_j(\phi_0) \neq 0 \}, \quad E := \operatorname{span}\{ e_j : j \in S \}.
\]

Since $\dim E \ge d+1$ and $\dim T = d$, we have
\[
\dim(E \cap T^\perp) \ge \dim E + \dim T^\perp - D \ge 1.
\]

Choose a unit vector $n \in E \cap T^\perp$.

Since $Y_j(\phi_0) \neq 0$ for $j \in S$, by continuity there exists a neighborhood $V \ni \phi_0$ such that $Y_j(\phi) \neq 0$ on $V$ for all $j \in S$. Choose $\chi \in C_c^\infty(V)$ with $\chi \equiv 1$ near $\phi_0$.

Define
\[
q_\varepsilon(\phi) := \frac{\varepsilon}{2}\|\phi - \phi_0\|^2.
\]

Define
\[
(h_\varepsilon)_j(\phi) =
\begin{cases}
\chi(\phi)\, q_\varepsilon(\phi)\, \dfrac{n_j}{Y_j(\phi)}, & j \in S,\\
0, & j \notin S.
\end{cases}
\]

Set $\tilde g_\varepsilon = g + h_\varepsilon$, $\tilde F_\varepsilon = \mu_{\tilde g_\varepsilon}$.

Then
\[
\tilde F_\varepsilon = F + \chi q_\varepsilon n.
\]

Since $q_\varepsilon$ vanishes to first order at $\phi_0$, we have
\[
D(\chi q_\varepsilon n)(\phi_0) = 0,
\]
so
\[
D\tilde F_\varepsilon(\phi_0) = DF(\phi_0).
\]

Thus $\tilde g_\varepsilon \in \mathcal I_{\phi_0}$.

\textbf{Second fundamental form.}
At $\phi_0$, choose normal coordinates on the parameter domain and orthonormal bases of $T$ and $T^\perp$, with $n = n^1$. In these coordinates, the Christoffel symbols vanish at $\phi_0$, so
\[
B^\alpha_{ij} = \langle \partial_{ij}F(\phi_0), n^\alpha \rangle.
\]

Since $\chi \equiv 1$ near $\phi_0$,
\[
D^2\tilde F_\varepsilon(\phi_0) = D^2F(\phi_0) + \varepsilon I_d \otimes n.
\]

Thus
\[
\tilde B^1_{ij} = B^1_{ij} + \varepsilon \delta_{ij}, 
\quad
\tilde B^\alpha_{ij} = B^\alpha_{ij} \ (\alpha \ge 2).
\]

\textbf{Curvature computation.}
By the Gauss equation,
\[
R_{ijkl} = \sum_\alpha (B^\alpha_{ik}B^\alpha_{jl} - B^\alpha_{il}B^\alpha_{jk}).
\]

Fix any $i \neq j$ (which exists since $d \ge 2$). Then
\[
\tilde R_{ijij}(\varepsilon)
=
(B^1_{ii}+\varepsilon)(B^1_{jj}+\varepsilon) - (B^1_{ij})^2
+ \sum_{\alpha \ge 2} (B^\alpha_{ii}B^\alpha_{jj} - (B^\alpha_{ij})^2).
\]

Hence
\[
\tilde R_{ijij}(\varepsilon)
=
R_{ijij}(0)
+ \varepsilon(B^1_{ii}+B^1_{jj})
+ \varepsilon^2.
\]

Since $n$ is a unit vector, the coefficient of the $\varepsilon^2$ term is exactly $1$. Thus this is a nonzero polynomial in $\varepsilon$, so it has only finitely many roots.

Therefore, for all sufficiently small $\varepsilon$ outside this finite set,
\[
\tilde R_{ijij}(\varepsilon) \neq 0.
\]

Since a single nonzero curvature component implies $R_{\tilde g_\varepsilon}(\phi_0) \neq 0$, we obtain $\tilde g_\varepsilon \in \mathcal N_{\phi_0}$.

Since $h_\varepsilon \to 0$ in $C^2$, every neighborhood of $g$ contains such perturbations. Thus $\mathcal N_{\phi_0}$ is dense in $\mathcal I_{\phi_0}$.
\end{proof}

\newpage

\subsection{Additional analysis of anisotropy under varying conditioning.}

\begin{figure}[H]
    \centering
    \includegraphics[width=0.75\linewidth]{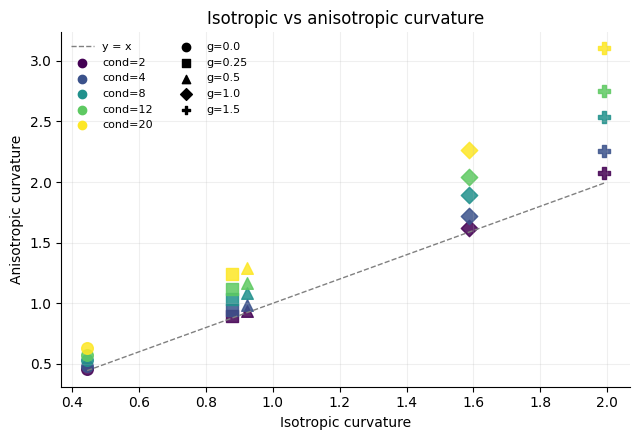}
    \caption{\textbf{Isotropic vs anisotropic curvature.} Points correspond to different gate strengths and condition numbers, with marker shape indicating gate strength and color indicating condition number. The two curvature measures are nearly perfectly correlated, while anisotropic curvature differs in scale due to metric effects.}
    \label{fig:iso_vs_aniso}
\end{figure}

\paragraph{Curvature is intrinsic to the representation mapping.}
Figure~\ref{fig:iso_vs_aniso} compares curvature measured under isotropic and anisotropic metrics across gate strengths and condition numbers. Each point corresponds to a specific combination of gate strength and condition number. We observe that isotropic and anisotropic curvature are nearly perfectly correlated across all settings. While anisotropic curvature increases with the condition number due to metric scaling, the relative ordering induced by gate strength is preserved. This shows that curvature is primarily determined by the representation mapping, while the choice of metric mainly affects its scale, consistent with our theoretical analysis.

\subsection{Ablation under anisotropic metrics}
\label{sec:appendix_ablation_aniso}

\begin{figure}[H]
    \centering
    \includegraphics[width=0.7\linewidth]{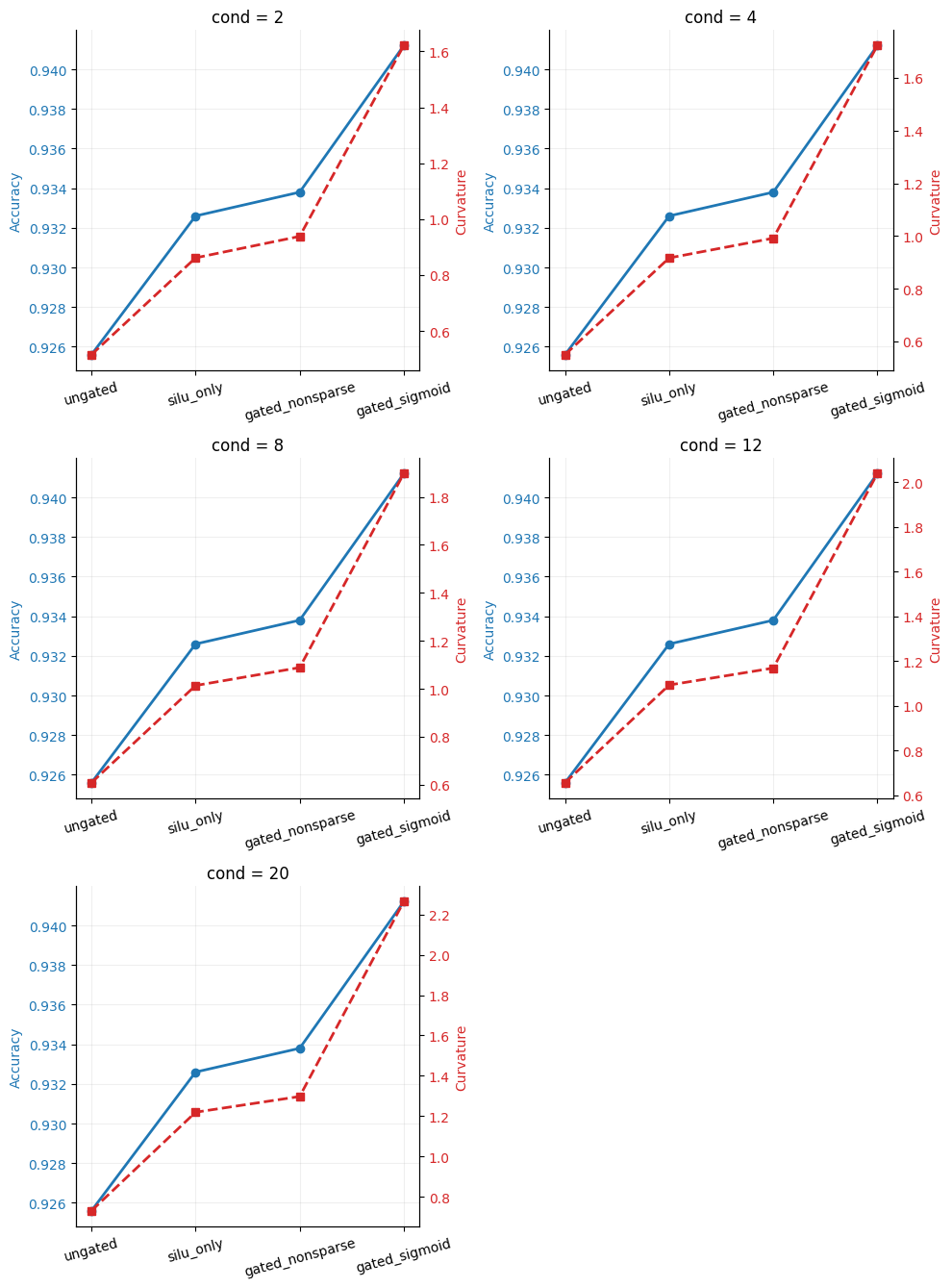}
    \caption{\textbf{Ablation under anisotropic metrics.} Each subplot corresponds to a different condition number. While curvature increases with the condition number, the relative ordering of variants and their accuracy remain unchanged.}
    \label{fig:ablation_aniso}
\end{figure}

We repeat the ablation analysis under anisotropic metrics with varying condition numbers. Figure~\ref{fig:ablation_aniso} shows test accuracy and anisotropic curvature for different attention variants across condition numbers.

We observe that the qualitative trends remain unchanged. Multiplicative gating consistently yields higher curvature and improved accuracy, while ungated attention and pointwise nonlinearities exhibit lower curvature and weaker performance. Increasing the condition number rescales the magnitude of curvature but does not affect accuracy or the relative ordering of variants. These results further support that geometric expressivity is driven by the representation mapping rather than the choice of metric.

\subsection{Linear Control Task}

To assess whether the observed gains arise from generic additional nonlinearity, we consider a control task in which the decision boundary is linear and does not require curved representations. As in the main experiment, each example is generated by sampling a latent center $c \in [-2,2]^2$ and forming a sequence of noisy observations. The label is determined by a linear function of the latent center,
\[
y = \mathbf{1}(w^\top c > 0),
\]
for a fixed vector $w \in \mathbb{R}^2$.

Table~\ref{tab:linear_control} reports results across different gating strengths. We observe that all variants achieve similar accuracy, with differences well within the standard deviation across seeds. In particular, accuracy remains nearly constant despite substantial variation in representation curvature.

\begin{figure}[H]
    \centering
    \includegraphics[width=0.5\linewidth]{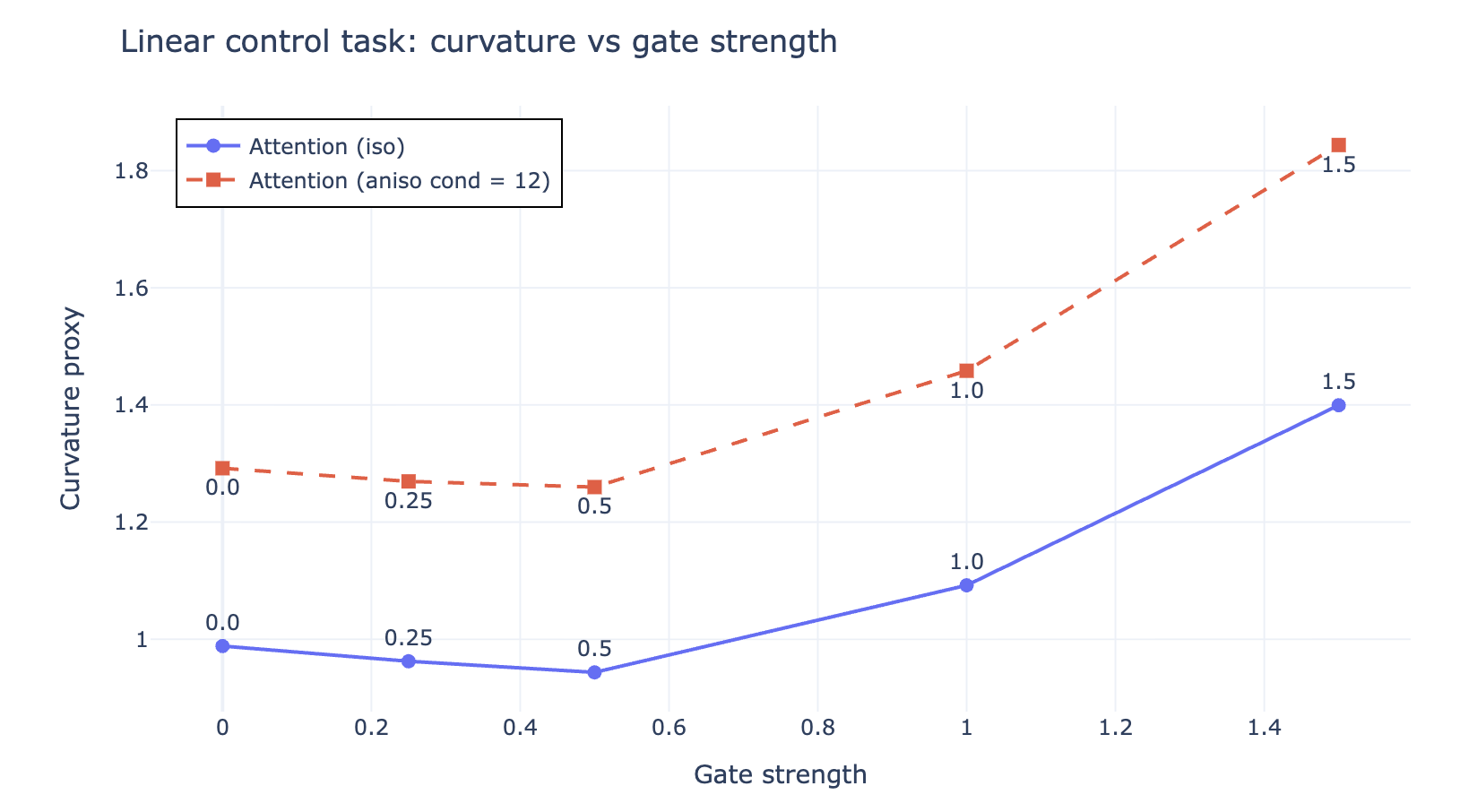}
    \caption{Linear control task. Attention curvature as a function of gate strength under isotropic and anisotropic metrics (condition number 12). While curvature increases overall with gate strength, the relationship is not strictly monotonic, reflecting the absence of task pressure for nonlinear structure.}
    \label{fig:linear_control_curvature}
\end{figure}

\begin{figure}[H]
    \centering
    \includegraphics[width=0.5\linewidth]{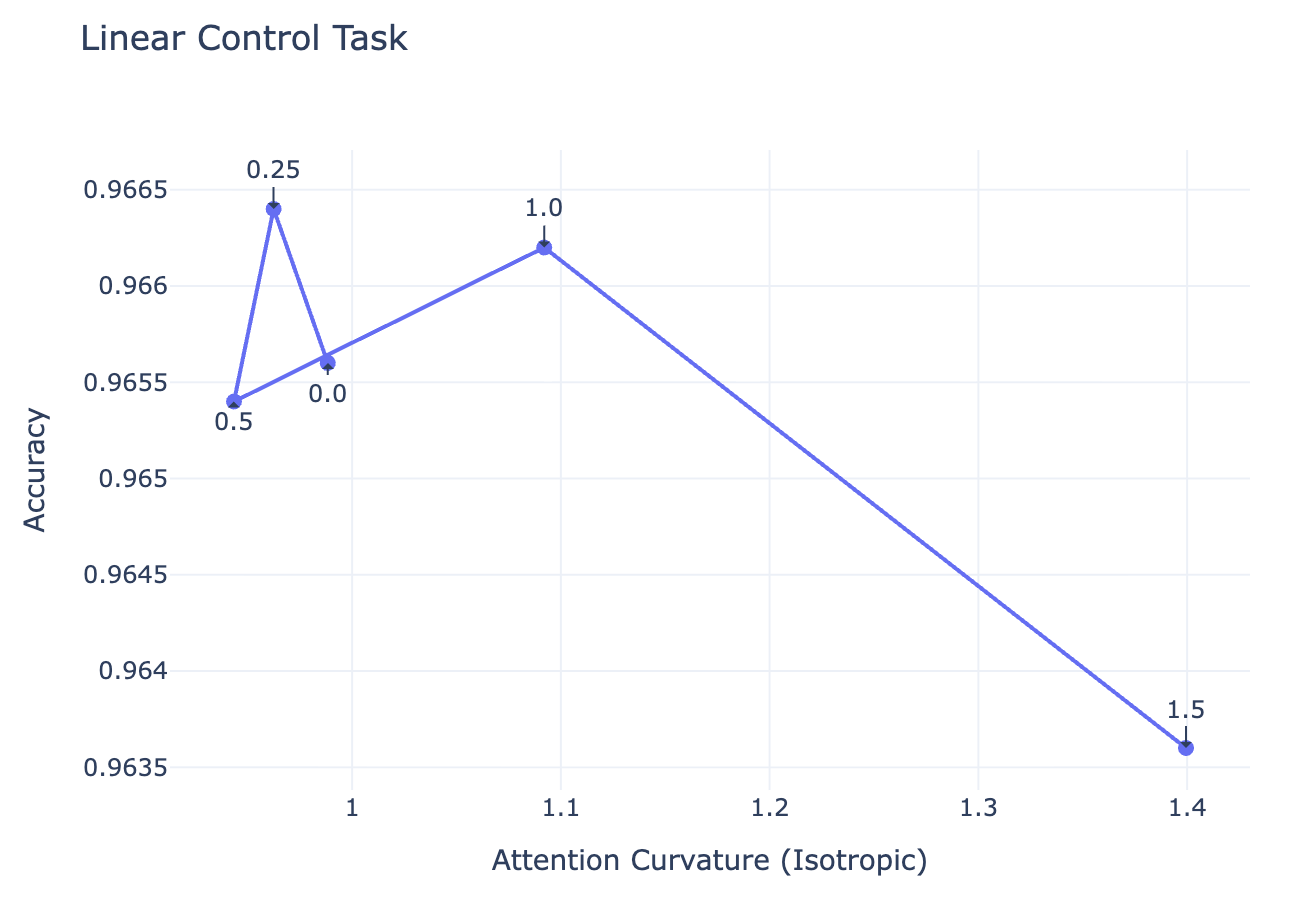}
    \caption{Linear control task. Accuracy as a function of isotropic attention curvature across gate strengths. No consistent positive relationship is observed, indicating that increased curvature does not improve performance when the task does not require nonlinear structure.}
    \label{fig:linear_control_accuracy}
\end{figure}

Figure~\ref{fig:linear_control_curvature} shows that increasing gate strength leads to a clear increase in attention curvature under both isotropic and anisotropic metrics (with condition number 12). However, as shown in Figure~\ref{fig:linear_control_accuracy}, these increases in curvature do not translate into improved performance. The relationship between curvature and accuracy is not consistently positive and is weakly negative in this setting.

Taken together, these results indicate that increased geometric expressivity does not provide a performance benefit when the task does not require nonlinear structure. This supports the interpretation that the gains observed in the main experiment arise specifically from the ability of gated attention to realize curved representations, rather than from generic additional nonlinearity.

\begin{table}[h]
\centering
\caption{Linear control task. Mean $\pm$ std over seeds.}
\label{tab:linear_control}
\begin{tabular}{c c c c}
\toprule
Gate strength & Accuracy & Attn curvature (iso) & Attn curvature (aniso cond = 12) \\
\midrule
0.00 & 0.9656 $\pm$ 0.0081 & 0.9883 & 1.2919 \\
0.25 & 0.9664 $\pm$ 0.0079 & 0.9624 & 1.2695 \\
0.50 & 0.9654 $\pm$ 0.0081 & 0.9434 & 1.2598 \\
1.00 & 0.9662 $\pm$ 0.0082 & 1.0920 & 1.4584 \\
1.50 & 0.9636 $\pm$ 0.0088 & 1.3995 & 1.8438 \\
\bottomrule
\end{tabular}
\end{table}

We note that, unlike in the curved task, curvature does not increase monotonically with gate strength in the linear control setting. This suggests that, in the absence of task pressure for nonlinear structure, curvature is not systematically reinforced during training.

\subsection{Experimental Details}

\subsubsection{Model Architecture}

We use a minimal attention-based model consisting of a single attention block followed by a classifier head. Inputs in $\mathbb{R}^2$ are projected to a hidden dimension of 64 and arranged into sequences of length 8. The projected inputs are processed by scaled dot-product attention, followed by mean pooling and a two-layer MLP classifier.

We consider multiple variants of the attention output:
(i) ungated attention,
(ii) a pointwise SiLU nonlinearity applied to the attention output, and
(iii) multiplicative gating with different parameterizations.

For gated models, we vary a gate strength parameter $\alpha \in \{0, 0.25, 0.5, 1.0, 1.5\}$, which controls the degree of multiplicative modulation.

\subsubsection{Training Setup}

All models are trained using AdamW with learning rate $2 \times 10^{-3}$, weight decay $10^{-4}$, and batch size 128 for 20 epochs. Results are averaged over 5 random seeds $\{0,1,2,3,4\}$. All model variants are trained under identical settings to ensure fair comparison.

\subsubsection{Data Generation}

We evaluate models on a synthetic sequence classification task designed to require nonlinear structure. Each dataset consists of 4000 training samples and 1000 test samples. Inputs are generated in $\mathbb{R}^2$ with latent centers sampled uniformly from $[-2,2]^2$, and observation noise with standard deviation 0.20 is added to each sample.

\subsubsection{Curvature Estimation}

We estimate representation curvature using a finite-difference proxy:
\[
\kappa(x) = \left\| \frac{f(x + \varepsilon v) - 2f(x) + f(x - \varepsilon v)}{\varepsilon^2} \right\|,
\]
where $v$ is a random unit direction and $\varepsilon = 10^{-2}$. The estimate is averaged over 64 random directions.

This quantity measures the magnitude of second-order variation of the representation map $f$ and serves as a proxy for local nonlinearity. It is not an invariant notion of Riemannian curvature and does not directly estimate sectional curvature. Rather, it is used for comparative analysis across model variants.

We report curvature of the attention output mean, and additionally compute curvature under a Fisher--Rao square-root embedding.

\subsubsection{Robustness Under Anisotropic Metrics}

To assess robustness under different metric scalings, we evaluate curvature under diagonal precision matrices with condition numbers $\{2, 4, 8, 12, 20\}$. This tests whether relative curvature comparisons are stable under anisotropic rescaling of the representation space.

\end{document}